\documentclass[sigconf]{acmart}

\AtBeginDocument{%
  }

\usepackage{algorithm}
\usepackage{algpseudocode}
\usepackage{natbib}  
\usepackage{caption} 
\usepackage{amstext}
\usepackage[switch]{lineno}
\usepackage{bm}
\usepackage{subcaption}
\usepackage{graphicx}
\usepackage{multirow}
\usepackage{enumitem}
\usepackage{tikz}
\usetikzlibrary{trees}
\usetikzlibrary{shapes,arrows}
\usepackage{adjustbox}
\usepackage{cleveref}
\usepackage{xcolor}
\tikzstyle{block} = [draw, fill=white, rectangle, minimum height=2em, minimum width=4em]
\tikzstyle{sum} = [draw, fill=white, circle, scale=0.005, node distance=0.5cm]
\tikzstyle{input} = [coordinate]
\tikzstyle{output} = [coordinate]  
\tikzstyle{pinstyle} = [pin edge={to-,thin,black}]
\newcommand{\ie}{\textit{i}.\textit{e}., }
\usepackage[autostyle]{csquotes}
\setcopyright{none}
\renewcommand\footnotetextcopyrightpermission[1]{}
\begin{document}

\title{EMCNet : Graph-Nets for Electron Micrographs Classification}

\author{Sakhinana Sagar Srinivas}
\authornote{Both authors contributed equally to this research.}
\email{sagar.sakhinana@tcs.com}
\affiliation{%
  \institution{TCS Research}
  \city{Bangalore}
  \country{India}
  \postcode{411013}
}

\author{Rajat Kumar Sarkar}
\authornotemark[1]
\email{rajat.sarkar1@tcs.com}
\affiliation{%
  \institution{TCS Research}
  \city{Pune}
  \country{India}
  \postcode{411013}
}

\author{Venkataramana Runkana}
\email{venkat.runkana@tcs.com}
\affiliation{%
  \institution{TCS Research}
  \city{Pune}
  \country{India}
  \postcode{411013}
}

\renewcommand{\shortauthors}{Sagar and Rajat et.al}


\begin{abstract}

Characterization of materials via electron micrographs is an important and challenging task in several materials processing industries. Classification of electron micrographs is complex due to the high intra-class dissimilarity, high inter-class similarity, and multi-spatial scales of patterns. However, existing methods are ineffective in learning complex image patterns. We propose an effective end-to-end electron micrograph representation learning-based framework for nanomaterial identification to overcome the challenges. We demonstrate that our framework outperforms the popular baselines on the open-source datasets in nanomaterials-based identification tasks. The ablation studies are reported in great detail to support the efficacy of our approach.
\vspace{-2mm}
\end{abstract}


\maketitle

\vspace{-2mm}
\section{Introduction}

Quality control in materials processing is of paramount importance across diverse industrial sectors such as batteries, semiconductors, ~\emph{etc}. Electron microscopes are used to create high-resolution images of materials known as electron micrographs. They provide the topography, morphology, composition, and crystallographic information to examine the structure of materials at the nanoscale. The nanomaterial recognition tasks\cite{aversa2018first} are challenging compared to the image-recognition benchmark datasets in degree of detail, complexity of patterns, and information density. The conventional modeling approaches for image-recognition tasks such as ConvNets(\cite{iandola2016squeezenet}, \cite{he2016deep}), Vision transformers(ViTs, \cite{dosovitskiy2020image}, \cite{SwinT}, \cite{ConViT}, \cite{Crossvit}, \cite{T2TViT}, \cite{ViT-SD}) and hybrid architectures(\cite{CVT}, \cite{Levit}, \cite{PVT}) suffer from inherent drawbacks of (a) requirement of large training datasets to introduce appropriate inductive biases and (b) large scale model complexity. The traditional Graph Neural Networks(GNNs, \cite{li2015gated}, \cite{morris2019weisfeiler}, \cite{velivckovic2017graph}, \cite{klicpera2018predict}, \cite{fey2019just}, \cite{chen2020}) learn and encode the complex discrete graph data by summarizing the high-level feature information in the graph-level embedding. The graph representation learning presents an alternate paradigm of approach for automatic identification of nanomaterials from their morphology or shape in the electron micrographs. The graph representation of the complex nanomaterial images has the spatial hierarchies of the high-level visual features embedded in a wide spectrum of graph structural-property information spanning across nodes, edges, motifs, and subgraphs. There is a need and necessity to tailor the architecture of traditional GNNs to effectively learn the structural characteristics of the graph for improved performance in the graph-level classification tasks. This work presents an overarching view of electron micrographs by effectively learning the multilevel and multiview representations for the global reasoning of the complex visual content embedded in the graph. Contrary to traditional CNNs, the novel framework identifies the discrete entities (low-level visual elements) and their pairwise relationships from the fixed-graph topology and learns high-level visual elements for better visual perception of nanostructures in the images. The proposed framework offers better generalization and scalability for large datasets.

\vspace{-2mm}
\section{Our Approach}
\begin{figure}[htbp]
\centering

\begin{tikzpicture}[auto, node distance=3cm,>=latex']
    \node [input, name=input] {};
    \node [sum, right of=input] (sum) {};
    \node [block,  below left of=sum, node distance = 1.0cm, xshift=-0.8cm, yshift=0cm] (controller) {$\textbf{GEnc}$};
    \node [block, below of=sum, node distance = 1.0cm, yshift=0.2cm] (system) {$\textbf{HGEnc}$};
    \node [block, below right of=sum, node distance = 1.0cm, xshift=0.8cm, yshift=0cm] (system_) {$\textbf{CTEnc}$};
    \node [block, below of=sum, node distance = 2.0cm] (system__) {$\textbf{Output Layer}$};

    \draw [->] (sum) -- node[label={[xshift=0.3cm, yshift=0.15cm]$\textbf{Input Graph}$}] {} (controller);
    \draw [->] (sum) -- node[label={[xshift=-1.0cm, yshift=0.15cm]$ $}] {} (system);
    \draw [->] (sum) -- node[label={[xshift=-1.0cm, yshift=0.15cm]$ $}] {} (system_);

    \node [output, below of=system, node distance = 0.85cm] (output) {};
    \draw [->] (system) -- node[label={[xshift=0.25cm, yshift=-0.45cm]}] {} (output); 
    
    \node [output, below right of=controller, xshift=-0.65cm, yshift=0.15cm, node distance = 1.62cm] (output) {};
    \draw [->] (controller) -- node[label={[xshift=-0.5cm, yshift=-0.5cm]}] {} (output); 

    \node [output, below left of=system_, xshift=0.65cm, yshift=0.15cm, node distance = 1.62cm] (output) {};
    \draw [->] (system_) -- node[label={[xshift=0.3cm, yshift=-0.35cm]}] {} (output); 
    
    \node [output, below of=system__, node distance = 0.8cm] (output) {};
    \draw [->] (system__) -- node[label={[xshift=0.6cm, yshift=-0.35cm]$\textbf{Label}$}] {} (output); 

\end{tikzpicture}
\vspace{-2mm}
\caption{\small{Overview of EMCNet framework, (a) we represent each image as a grid graph, (b) the $\textbf{GEnc}$ and $\textbf{HGEnc}$ modules of the framework  computes the respective grid graph representations, (c) the $\textbf{CTEnc}$ module determines the tree representation, and (d) the output layer predicts the image category.}} \label{fig:fig6}
\end{figure}
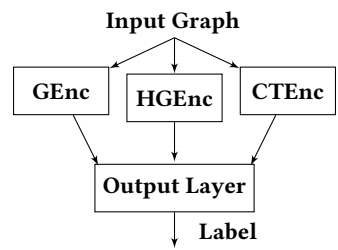

This section presents the Electron Micrographs Classification Network for brevity, $\text{EMCNet}$ for nanomaterials identification. Our framework consists of six main steps, (1) tokenization of the nanoscale images; splitting each image into patches and representing it as a patch-attributed grid graph with diagonal edges, (2) Graph encoder, for brevity, $\textbf{GEnc}$; performs the iterative neighborhood-local aggregation schemes on the augmented graphs for learning the abstract graph representations, (3) Hierarchical graph encoder, for brevity, $\textbf{HGEnc}$; performs the layer-wise local-graph pooling and higher-order message-passing schemes to learn the implicit multi-grained hierarchical representations, (4) tree decomposition; we perform the tree decomposition of the grid graph to obtain a clique tree graph,(5) Clique tree encoder, for brevity, $\textbf{CTEnc}$; performs the iterative neighborhood message-passing schemes on the clique tree graph to learn the local substructures in the graph, and (6) the output layer; applies a weighted linear transformation on the learned representations and predicts the image category. The overview of the EMCNet framework is illustrated in Figure \ref{fig:fig6}. 

\vspace{-2mm}

\vspace{-6mm}
\subsection{Tokenization of Images}
\label{sec:TI} 

Let us assume that $\text{H}$, $\text{W}$, and $\text{C}$ denote the dimensions of an RGB image, where $\text{H}$ is the height, $\text{W}$ is the width, and $\text{C}$ is the number of channels. For a given 3D image, $\text{I} \in \mathbb{R}^{H\times W\times C}$ and the resolution of each patch $(\text{P}, \text{P}, \text{C})$. We split the 3D image into \enquote{$\mathbf{N}$} non-overlapping patches(also known as tokens). $\mathbf{N}$ is determined by $\frac{\text{HW}}{\text{P}^{2}}$. In a nutshell, each 3D image is reshaped to obtain a sequence of flattened 2D patches, $\text{I}_{P} \in \mathbb{R}^{N\times P^{2}C}$. The independent patches are projected into a $d$-dimensional embedding space to obtain the low-level patch representations, $\text{I}_{P} \in \mathbb{R}^{N\times d}$. It is described below,
\vspace{-.5mm}
\begin{equation}
\text{I}_{P} = \text{I}_{P}\mathbf{E} 
\end{equation}

\begin{figure}[htbp]
\hspace*{-1.1cm} 
    \includegraphics[trim={0 0cm 0cm 0}, width=0.55\textwidth]{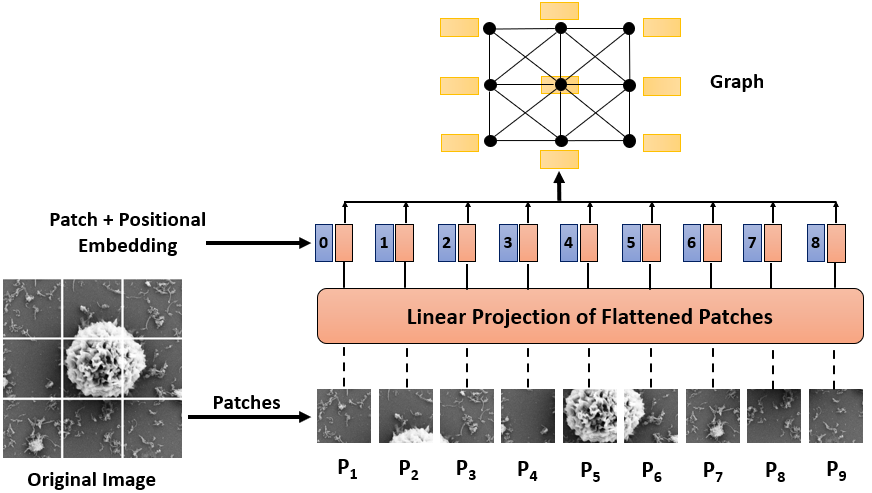}
    \vspace{-6mm}
    \caption{\small{For illustration purpose, we split the image into $\textbf{3}\times \textbf{3}$ patches and represent it as an undirected graph.}}
    \label{fig:fig1}
\end{figure}

\begin{figure}[htbp]
    \centering
    \includegraphics[trim={0 0cm 0cm 0},width=0.425\textwidth]{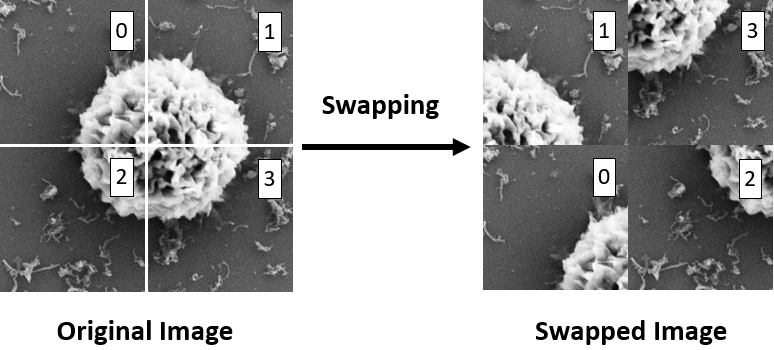}
    \vspace{-3mm}
    \caption{\small{For illustration purpose, we split an image into $\textbf{2}\times \textbf{2}$ patches. In the absence of positional information, the GNNs cannot distinguish the images on the left and right.}}
    \label{fig:fig2}
\end{figure}

The trainable patch embeddings, $\mathbf{E} \in \mathbb{R}^{P^{2}C\times d}$ are the semantic representations of the patches in vision tasks. The patch representations suffer from the inherent drawback of lack of patch locality-preserving information. The position embeddings, $\text{P}_{E} \in \mathbb{R}^{N\times d}$ are linearly added to the patch representations to enable position awareness, refer to Figure \ref{fig:fig1}. The position embeddings along with the patch embeddings are randomly initialized and are jointly updated along with the network parameters during the training of the model. We represent the patches of each image as the nodes of a regular static graph. In general, there is no natural ordering of the nodes in the graph. Here, we expose our model to the positional information of the patches to effectively exploit the locality information to perform better on the evaluation metrics. Figure \ref{fig:fig2} depicts the illustration of the tokenization of the images.

\vspace{-1mm}

\vspace{-2mm}

\vspace{-4mm}
\subsection{Graph Representation}
\label{sec:GR}

We represent each nanoscale image as the node-attributed undirected graph, $\mathcal{G}=(\mathcal{V}, \mathcal{E})$. The node set, $\mathcal{V}$ denote the sequence of patches. The node feature matrix is described by $X \in \mathbb{R}^{N\times d}$. $N(|\mathcal{V}|)$ denotes the number of nodes. $d$ is the dimensionality of the node feature vector. The patches are connected through an edge to adjacent patches. $\mathcal{E}$ denotes the edge-set. It is obtained from the graph structure information. The graph structure can also be represented by the adjacency matrix, $\mathcal{A} \in \mathbb{R}^{N\times N}$. $\mathcal{A}[v, u]=1$ if $(v,u) \in \mathcal{E}$ or else $\mathcal{A}[v, u]=0$. Here, $u$ \&  ${v \in \mathcal{V}}$ refers to the local-graph neighbors and are connected by an arbitrary edge characterized by an empty edge attribute, $(v,u) \in \mathcal{E} \leftrightarrow(u, v) \in \mathcal{E}$, $\forall v \in \mathcal{N}(u)$. The local-graph neighbors of each node are described by $\mathcal{N}(u) = \left\{v \in \mathcal{V} \mid (v,u) \in \mathcal{E} \right\}$. The graph representation of the nanoscale images has a grid-like structure and has fixed dependency relationships among the nodes of the graph across the diverse set of images. Given a dataset $(\mathcal{G}_{i}, y_{i})=\left\{\left(\mathcal{G}_{1}, y_{1}\right), \cdots, \left(\mathcal{G}_{n}, y_{n}\right)\right\}$, where $y_{i}$ is the ground truth label of $\mathcal{G}_{i}$, the objective of the graph-level classification task is to learn a novel mapping function $f: \mathcal{G}_{i} \rightarrow y_{i}$ that maps the discrete graphs to the set of labels.

\vspace{-4mm}
\subsection{Graph Encoder(GEnc)}
\label{sec:GEnc}

The objective of the graph encoder is to map the high-dimensional discrete graphs information to the low-dimensional graph-level representations in the Euclidean space. 
We propose a message-passing neural network to model the discrete graphs, $\mathcal{G}_i$ for determining both the node-level $\mathbf{z}_{u}, u \in \mathcal{V}(\mathcal{G}_i)$, and the graph-level abstract representations $\mathbf{z}_{\mathcal{G}_i}$, whilst maximally preserving the high-level visual feature information embedded in the graphs. We augment each graph with a virtual master node. The master node is bidirectionally connected to all nodes of the graph through virtual edges. The virtual edges represent the pairwise relationships between each \enquote{real} node and the master node in the graph. The augmented graph is not a fully-connected graph. The virtual master node representation contains the global-information of the nodes. Each node reads and writes to transform the virtual master-node representation through the iterative message-passing schemes. Neural messages are communicated across the nodes of the augmented graphs through the edges connecting them. The neural messages encapsulate the immediate graph neighbors information. The neighborhood-local aggregation schemes learn the node-level abstract representations. The node representations capture the rich local and informative long-range pairwise interactions spanning across the graph. The virtual master node helps in providing an additional path for message propagation to promote the information diffusion between the distant nodes of the graph. It helps in learning the global structural properties for a graph-wide representation. Each node in the graph receives and sends neural messages to its local-graph neighbors. At first, the source node $v \in \mathcal{V}$ receives neural messages $\Phi_{w v}$ from its local-graph neighbors, $w \in \mathcal{N}(v) \backslash u$. $\Phi_{w v}$ is modeled by a linear operator to encode the neighboring node's abstract representations. The subscript, $wv$ indicates the message propagation from the node $w$ to node $v$. Subsequently, the source node, $v \in \mathcal{N}(u)$ modifies the neural message vectors $\Phi_{w v}$ with its feature vector $x_{v}$, and applies non-linearity to compute the neural-message vector $\Phi_{vu}$. The source node sends the neural message $\Phi_{vu}$ to the sink node $u$ to refine the node embedding $\mathbf{z}_{u}$. In short, each source node, $v \in \mathcal{V}$ dispatches neural messages to the sink node, $u \in \mathcal{V}$ only after it receives all the incoming neural messages from its neighbors, $\Phi_{w v}, w \in \mathcal{N}(v) \backslash u$. Algorithm \ref{alg:1} presents our Graph Encoder(GEnc). For each message-passing iterative step, the receptive field increases by one hop. We compute the refined neural message vector by instigating additional rounds of message-passing to embed the larger neighborhood of each node by iterating for T steps. Each node in the graph perceives the refined neural messages sent from its immediate graph neighbors to transform its abstract representation. The transformed node representations encode the local substructure information about its T-hop neighborhood. We determine the fixed-size graph-level embedding $\mathbf{z}_{\mathcal{G}_i}$ by performing the sum-pooling operation on the node-level embeddings. The weight matrix, $\text{W}^{g}$ applies a shared linear transformation on each node feature vector. The trainable parameters, $\text{U}^{g}_{1}$ and $\text{U}^{g}_{2}$ of the iterative graph message-passing schemes are shared across nodes of the graph. $\tau$ is a non-linear sigmoidal function. The optimal selection of message-passing iterative-steps (neighborhood aggregation) trade-offs the under-representation and the over-smoothing of the node-level representations of the augmented graph to encode the larger local-graph neighborhood. The graph-level representation obtained from the node-level representations of the augmented graphs with fixed graph topology and varying node features are discriminative enough to achieve higher classification accuracy. In summary, the graph encoder maps the discrete graph data to the optimal graph-level representations by learning to model the visual information embedded in the complex structural properties of the graph. It is utilized in the downstream tasks to perform inference on the graphs through label predictions. 

\vspace{-2mm}
\begin{algorithm}[htbp]
\caption{Iterative Graph Message Passing Mechanism}
\begin{algorithmic}
\normalsize

\State Initialize messages: $\Phi_{wv}^{(0)}=0$
\For{$t=1$ \textbf{to} $T$}  \\
    \state \hspace{5mm} $\Phi_{w v}^{(t)} = \tau \big(\text{W}^{g} x_{w} + \sum_{w \in  \mathcal{N}(v) \backslash u} \Phi_{w v}^{(t-1)}\big)$ \\
    \state \hspace{5mm} $\Phi_{vu}^{(t)}= \tau\big(\text{W}^{g} x_{v} + \Phi_{w v}^{(t)}\big)$ \\
  
\EndFor \\
\state  $\mathbf{z}_{u}=\tau\big(\text{U}^{g}_{1} x_{u} + \sum_{v \in \mathcal{N}(u)} \text{U}^{g}_{2} \Phi_{vu}^{(T)}\big)$ \\
\state  \textbf{Return} $\mathbf{z}_{\mathcal{G}_i}=\sum_{u} \mathbf{z}_{u} /|\mathcal{V}|$.
\end{algorithmic}
\label{alg:1}
\end{algorithm}

\vspace{-4mm}
\subsection{Hierarchical Graph Encoder(HG-Enc)}
\label{sec:HG-Enc}

\vspace{-2mm}
\begin{figure}[htbp]
    \centering
    \includegraphics[scale=0.27]{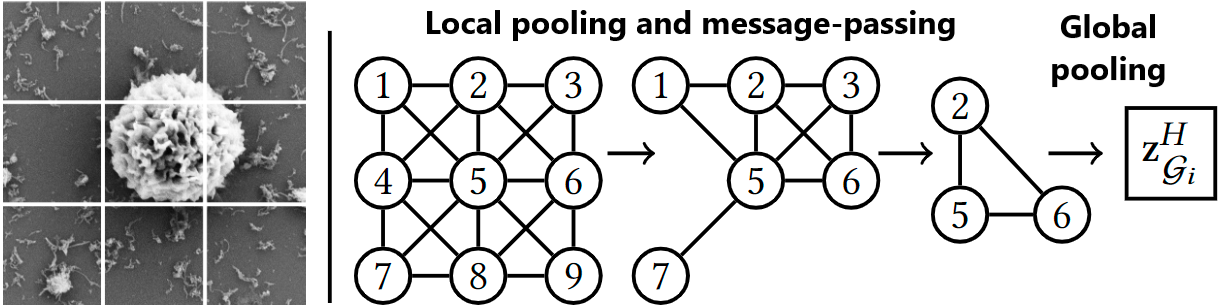}
    \vspace{-5mm}
    \caption{\small{The nodes are labeled for illustration. We gradually reduce the graph size in progressive layers, by rejecting the nodes of lower importance and learn the high-level representations through the self-attention based message-passing schemes.}}
    \label{fig:fig4}
\end{figure}

\vspace{-3mm}
The HG-Enc module learns the prominent subgraph structures to determine the hierarchical representations of the graphs. The HG-Enc module overcomes the following shortcomings, (1) traditional GNNs treat all the nodes equally by disregarding the subset of nodes of high importance; explicitly, it models all the patches of an image without consideration of the visual content, (2) the tokenization of the images results in non-overlapping patches and this skips the boundary-level information present in the adjacent patches of an image, and (3) there is a need to learn the scale-variant visual elements in the underlying graph, and this would be intractable for the conventional GNNs to encode from the independent patches of a fixed scale. Figure \ref{fig:fig4} depicts the illustration of the hierarchical graph encoder. The key steps in the HG-Enc module are (a) the basis for pooling the graph to obtain the pooled graph, (b) structure learning of the pooled graph, and (c) determining the higher-order node representations of the pooled graph. The HG-Enc module performs the non-linear down-sampling by adaptively rejecting a subset of nodes in the main graph to form an induced subgraph through an importance score measure. The importance score of each node is determined by performing a scalar projection of the node-level embedding on the trainable projection vector. The node embeddings are determined through the spatial-graph filtering operations.

\vspace{-4mm}
\begin{equation}
\normalsize
z^{(l+1)}_{u}=\operatorname{ReLU}\big(\alpha_{u, u} W^{(l)}z^{(l)}_{u} + \hspace{-2mm} \sum_{v \in \mathcal{N}^{(l)}(u)} \alpha_{v, u} W^{(l)}z^{(l)}_{v}\big)
\end{equation}
\vspace{-2mm}

where $z^{0}_{u} = x_{u} \in \mathbb{R}^{d}$ is the feature vector of node $u$. $z^{(l)}_{u} \in \mathbb{R}^{d}$ denotes the abstract representation of node $u$. 
The superscript, $l$ denotes the layer. The set of neighbors of a node $u$ obtained from the adjacency matrix $\mathcal{A}^{(l)}$, is $\mathcal{N}^{(l)}(u)=$ $\left\{v \mid \mathcal{A}^{(l)}[v,u]>0\right\}$. The trainable weight matrix, $\text{W}^{(l)} \in \mathbb{R}^{2d\times d}$ applies a shared linear transformation on the node feature vector, and the attention coefficients $\alpha_{v, u}$ are determined as,

\vspace{-5mm} 
\begin{align}
\normalsize
\zeta(v, u) &=\operatorname{ReLU}\left(a^{\top}\left(r_{u} \oplus r_{v}\right)\right); r_{u} = Wz^{(l)}_{u} \\
\alpha_{v, u} &=\frac{\exp (\zeta(v, u))}{\sum_{k \in \mathcal{N}^{(l)}(u) \cup\{u\}} \exp (\zeta(k, u))}
\end{align}

Where $a$ is the vector of the weighted coefficients for the attention mechanism. $\oplus$ denotes the concatenation operation. $r_{u}$ is the linearly transformed patch representations. $\operatorname{ReLU}$ is the non-linear activation function. The layer-wise forward propagation of the HG-Enc operator is described by,

\vspace{-5mm}
\begin{align}
\normalsize
\hspace{10mm} s_{u} =  z^{(l+1)}_{u}\frac{p^{(l+1)}}{ \lVert p^{(l+1)} \rVert}  
\end{align}

 where $z^{(l+1)}_{u}, p^{(l+1)} \in \mathbb{R}^{2d}$ \& $s \in \mathbb{R}^{|\mathcal{V}^{(l)}|}$. $p^{(l+1)}$ denotes the trainable projection vector to map the node representations into the importance score. The scalar projection of the node-attributes on the normalized projection vector is described by $s$. The importance score is utilized to rank the nodes in the graph to reduce the graph size. For a given pooling ratio $p_r$, we select a subset of top \hspace{0.5mm}$\text{m}$-ranked nodes in the main graph. The pooling ratio is described by, $p_r = \frac{m}{|\mathcal{V}^{(l)}|}$, $\forall$ $p_r \in (0,1]$. The node ranking operation, $\textit{idx} = \mbox{rank}(s, m)$ utilizes the scalar projection scores to sample the top \hspace{0.5mm}$\textit{m}$-prominent nodes. $\textit{idx}$ describes the indices of the selected top \hspace{0.5mm}$\textit{m}$-nodes in the graph. We obtain the pooled graph $\mathcal{G}^{(l+1)}$ from the main graph $\mathcal{G}^{(l)}$ by utilizing the prominent nodes indices obtained through the node-ranking operation. The pooled graph-structure and node attribute information is described as, 

\vspace{-4mm}
\begin{equation}
\mathcal{A}^{(l+1)} = \mathcal{A}^{(l)}(\textit{idx}, \textit{idx}); \hspace{0.5mm} \mathcal{A}^{(l+1)} \in \mathbb{R}^{m\times m}, \hspace{0.5mm} \mathcal{A}^{(l)} \in \mathbb{R}^{|\mathcal{V}^{(l)}|\times |\mathcal{V}^{(l)}|} \nonumber
\end{equation}

where $\mathcal{A}^{(l+1)}$ is the adjacency matrix of the pooled graph. The node attribute matrix of the pooled graph $\mathcal{G}^{(l+1)}$ is described by, 

\vspace{-3mm}
\begin{equation}
\tilde{Z}^{(l+1)} = Z^{(l)}(\textit{idx}, :); \hspace{0.5mm} \tilde{Z}^{(l+1)} \in \mathbb{R}^{m \times 2d}, \hspace{0.5mm} Z^{(l)} \in \mathbb{R}^{|\mathcal{V}^{(l)}| \times 2d}
\end{equation}

A gating operation is performed to regulate the information flow to avoid inadvertent over-smoothing of the node representations, 

\vspace{-3mm}
\begin{equation}
\tilde{s} = \mathrm{ReLU} \big(s(\textit{idx})\big) \label{eqn:five} 
\end{equation}
\vspace{-2mm}
\begin{equation}
Z^{(l+1)} = \tilde{Z}^{(l+1)} \odot \big(\tilde{s} \mathbf{1}^{T}_{2d}\big) \label{eqn:six}
\end{equation} 

$\mathbf 1_{2d}\in \mathbb{R}^{2d}$ is a vector of size $2d$ and with each component value of 1. $\odot$ denotes the element-wise product. We will apply the self-attention mechanism to learn the compact hierarchical node representations of the pooled graph from the input node representations as given by the node attribute matrix, $Z^{(l+1)}$, refer Equation \ref{eqn:six}. The hierarchical node representation is computed as the weighted sum of the node attributes of the pooled graph,

\vspace{-4mm}
\begin{equation}
z^{(l+1)}_{u}=\sum_{k} \alpha_{ku}\left(z^{(l+1)}_{k} W^{(l+1)}_{V}\right), k \in \mathcal{N}^{(l+1)}(u) \cup\{u\}
\end{equation}

where $z^{(l+1)}_{u} \in \mathbb{R}^{d}$. Each weight coefficient $\alpha_{ku}$ is computed using a softmax as,
\vspace{-4mm}
\begin{equation}
\alpha_{ku}=\frac{\exp \left(e_{ku}\right)}{\sum_{k} \exp \left(e_{ku}\right)}, 
\end{equation}

where $e_{vu}$ is calculated using scaled dot-product attention. It is described by,
\vspace{-2mm}
\hspace{5mm}\begin{equation}
e_{ku}=\frac{\big(z^{(l+1)}_{u} W^{(l+1)}_{Q}\big)\big(z^{(l+1)}_{k} W^{(l+1)}_{K}\big)^{T}}{\sqrt{d}}
\end{equation}

The trainable projections $W^{(l+1)}_{Q}, W^{(l+1)}_{K}, W^{(l+1)}_{V} \in \mathbb{R}^{2d \times d}$ are unique per layer. Here, we stack 3-layers of $\text{HG-Enc}$ operators each with a pooling ratio of $p_{r}=0.75$.  We perform the global average pooling operation on the node-level embeddings of the pooled graph to obtain $\mathbf{z}^{H}_{\mathcal{G}_i}$. There exist various techniques to perform down-sampling on graphs.
The local-graph pooling technique is fundamentally different from the graph coarsening(contraction) technique on the graph reduction method. The local-graph pooling technique ranks the nodes based on the complex visual content it contains. The low-ranking nodes are characterized by noise or contain the no-visual element. The HG-Enc operator drops the nodes of lower importance through the local-graph pooling technique and models the task-relevant induced subgraph (high-ranking nodes) from the main graph. In comparison, the graph coarsening technique\cite{ranjan2020asap} \cite{ying2018hierarchical} clusters the nodes into supernodes based on the learned node representations. It implicitly models the noise, learns sub-optimal node representations, and are less-effective on the graph-level classification tasks. In summary, the HG-Enc operator learns the long-range, multi-level dependencies of the graph.

\vspace{-2mm}
\subsection{Clique Tree Encoder(CTEnc)}

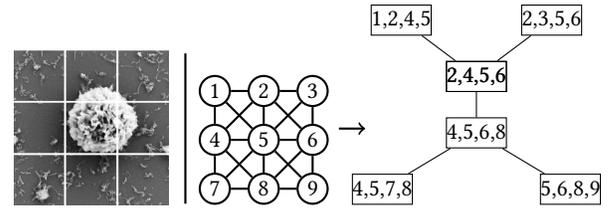
\begin{figure}[htbp]
\hspace{-1.5mm}\includegraphics[trim={0 0cm 0cm 0}, width=0.115\textwidth]{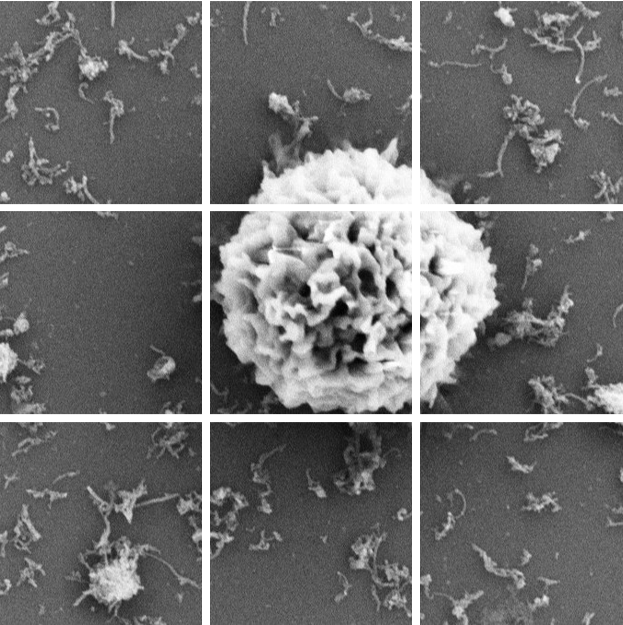}
\hspace{0.05cm}
\begin{tikzpicture}[node distance={5mm}, thick, main/.style = {draw, shape = circle, minimum size = 0.4cm, inner sep=0pt}] 
\draw[-, black, thick] (0.25,-1) -- (0.25,1);
\end{tikzpicture} 
\hspace{0.005cm}
\begin{tikzpicture}[node distance={6.5mm}, thick, main/.style = {draw, shape = circle, minimum size = 0.4cm, inner sep=0pt}] 
\node[main][] (1) {$1$}; 
\node[main][] (2) [right of=1] {$2$}; 
\node[main][] (3) [right of=2] {$3$}; 
\node[main][] (4) [below of=1] {$4$}; 
\node[main][] (5) [right of=4] {$5$}; 
\node[main][] (6) [right of=5] {$6$}; 
\node[main][] (7) [below of=4] {$7$}; 
\node[main][] (8) [right of=7] {$8$}; 
\node[main][] (9) [right of=8] {$9$}; 
\draw (1) -- (2); 
\draw (2) -- (3); 
\draw (1) -- (4); 
\draw (4) -- (7); 
\draw (2) -- (5); 
\draw (5) -- (8); 
\draw (3) -- (6); 
\draw (6) -- (9); 
\draw (4) -- (5); 
\draw (5) -- (6); 
\draw (7) -- (8); 
\draw (8) -- (9);
\draw (1) -- (5); 
\draw (5) -- (9); 
\draw (7) -- (5); 
\draw (5) -- (3);  
\draw (2) -- (4); 
\draw (4) -- (8); 
\draw (2) -- (6); 
\draw (8) -- (6); 
\draw[->, black, thick] (1.65,-0.5) -- (2,-0.5);   
\end{tikzpicture} 
\hspace{0.05cm}
\hspace{-4mm}\begin{tikzpicture}[nodes={draw, rectangle, minimum size = 0.4cm, inner sep=0pt}, -] 
\node at (2,0) {1,2,4,5} [sibling distance = 2.5cm, level distance = 0.75cm]
    child {node at (1,0) {2,4,5,6} 
    child {node at (0,0) {4,5,6,8}
    child {node at (0,0) {4,5,7,8}}
    child {node at (0,0) {5,6,8,9}}}
    };
\node at (4,0) {2,3,5,6} [sibling distance = 2.5cm, level distance = 0.75cm]
    child {node at (-1,0) {2,4,5,6} 
    };
\end{tikzpicture} 

\caption{\small{The nodes are labeled for illustration. We transform the graph into a hierarchical clique tree. The supernodes are known as cliques. The top-left-corner supernode of the clique tree is an induced subgraph(consists of nodes with labels 1, 2, 4, 5, and all of the edges) of the main graph and similarly the other supernodes.}}
\label{fig:fig5}
\end{figure}

We perform the tree-decomposition\cite{aguinaga2018learning} of the graph $\mathcal{G}$ to obtain the clique tree representation $\mathcal{T}$. It presents the tree-structured scaffold of the local graph substructures($\ie$ motifs and subgraphs) and their pairwise relationships. Figure \ref{fig:fig5} depicts an illustration of the tree decomposition of the graph representation of the nanoscale images. Each supernode of the clique tree, $\mathcal{C}_i = (\mathcal{V}_{i}, \mathcal{E}_{i})$ is an induced subgraph of the main graph, $\mathcal{G}=(\mathcal{V}, \mathcal{E})$. Each supernode of the clique tree, $C_{i}$ is labeled with a $\mathcal{V}_{i} \subseteq \mathcal{V}$ and $\mathcal{E}_{i} \subseteq \mathcal{E}$. Here, $i$ refers to the index of the supernode in $\mathcal{T}$. Each supernode, $\mathcal{C}_i$ in the clique tree $\mathcal{T}$ is described by a feature vector $x_{i}$. As stated earlier, each supernode in the clique tree is an induced subgraph of the main graph. We apply a shared linear transformation on the concatenated matrix of the node features in the subgraph to obtain the feature vector of the supernode. We utilize tree-based message-passing schemes to operate on the topology of the clique tree $\mathcal{T}$ to determine its low-level embedding $\mathbf{h}^{\mathcal{T}}$. A random leaf supernode is selected as the root supernode of the clique tree. Neural messages are computed and propagated across the supernodes of the clique tree in two phases. The neural messages encode the neighboring supernode's information in the clique tree and the local relations among the supernodes. At first, in the bottom-up phase, neural messages are propagated from the leaf supernodes iteratively towards the root supernode of the clique tree. In the top-down phase, neural messages are dispatched from the root supernode to its child supernodes down to the leaf supernodes of the clique tree. The neural messages $m_{ij}$ and $m_{ji}$ are dispatched from the supernode $\mathcal{C}_i$ to the supernode $\mathcal{C}_j$ through the superedge $(\mathcal{C}_i,\mathcal{C}_j)$ and the vice-versa in the clique tree. 

\vspace{-3mm}
\begin{algorithm}[ht]
\caption{Iterative Tree-Message Passing Mechanism}
\begin{algorithmic}
\normalsize

\State Initialize messages: $\mathbf{m}_{ji}^{(0)}=\mathbf{0}$
\For{$t=1$ \textbf{to} $T$} 
\State $\mathbf{m}^{(t)}_{ji}= \operatorname{GRU}\big(x_{j}, \big\{\mathbf{m}^{(t-1)}_{k j}\big\}_{k \in \mathcal{N}(j) \backslash i}\big)$. 
\EndFor
\State $\mathbf{h}_i=\tau\big(\mathbf{W}^{\mathcal{T}}_{1} x_{i} + \sum_{j \in \mathcal{N}(i)} \mathbf{W}^{\mathcal{T}}_{2} \mathbf{m}^{(T)}_{j i}\big)$.
\State \textbf{Return} $\mathbf{h}^{\mathcal{T}} = root \hspace{1mm} node, \mathbf{h}_i$.

\end{algorithmic}
\label{alg:2}
\end{algorithm}

\vspace{-6mm}
\begin{algorithm}[ht]
\caption{Gating Mechanism on Tree-Structure}
\begin{algorithmic}
\normalsize
\State $\mathbf{s}_{ji}=\sum_{k \in \mathcal{N}(j) \backslash i} \mathbf{m}_{k j}$ 
\State $\mathbf{z}_{ji}=\sigma\big(\mathbf{W}^{z} x_{j} + \mathbf{U}^{z} \mathbf{s}_{ji}+\mathbf{b}^{z}\big)$ 
\State $\mathbf{r}_{kj}=\sigma\big(\mathbf{W}^{r} x_{j} + \mathbf{U}^{r} \mathbf{m}_{kj}+\mathbf{b}^{r}\big)$  
\State \hspace{-1mm}$ \widetilde{\mathbf{m}}_{ji}=\tanh \big(\mathbf{W} x_{j}+\mathbf{U} \sum_{k \in \mathcal{N}(j) \backslash i} \mathbf{r}_{kj} \odot \mathbf{m}^{(t-1)}_{kj}\big)$ 
\State \hspace{-3mm}$\quad \mathbf{m}_{ji}=\big(1-\mathbf{z}_{kj}\big) \odot \mathbf{s}_{ji} + \mathbf{z}_{kj} \odot \widetilde{\mathbf{m}}_{kj}$

\end{algorithmic}
\label{alg:3}
\end{algorithm}

\vspace{-4mm}

\vspace{-3mm}
Algorithm \ref{alg:2} presents our Clique Tree Encoder(CTEnc). The neural message $\mathbf{m}_{ji}$ is computed through the gating mechanism to encapsulate the long-range interactions across the supernodes of the clique tree by regulating the information diffusion. Algorithm \ref{alg:3} presents the gating mechanism on the tree structure. After performing, the $\text{T}$-steps of message-passing iterations. We perform the sum-pooling operation on the weighted neural messages dispatched from the local-tree neighbors, $j \in \mathcal{N}(i)$ to compute a single-message vector. The supernode $\mathcal{C}_i$ perceives the aggregated message vector to refine its embedding $\mathbf{h}_i$. The transformed supernode embeddings of the clique tree incorporate information about their local $\text{T}$-hop neighbors. The tree-level embedding $\mathbf{h}^{\mathcal{T}}$ is the root-node embedding, $\mathbf{h}_{\text{root}}$ and it summarizes the subgraph patterns in the main graph. The trainable parameters of the learning algorithm $\mathbf{W}^{\mathcal{T}}_{1}$, $\mathbf{W}^{\mathcal{T}}_{2}$ are shared across the supernodes of the clique tree. $\tau$ is the sigmoidal function that introduces the non-linearity in the message-passing schemes. In summary, the tree encoder maps the discrete clique trees to determine the low-level tree representations. 

\vspace{-2mm}
\section{Output Layer}
\label{sec:PF} 

The outputs of the parallel operating modules, $\text{GEnc}$, $\text{HGEnc}$, and $\text{CTEnc}$ are further transformed by the linear operator. We apply softmax-activation on the output of the linear operator for the multi-class classification task. It is mathematically described as follows, 

\vspace{-5mm}
\begin{equation}
\textbf{q}_{i} = \textbf{softmax}\big( \textbf{W}_{1}\mathbf{z}_{\mathcal{G}_i} + \textbf{W}_{2}\mathbf{z}^{H}_{\mathcal{G}_i} + \textbf{W}_{3}\mathbf{h}^{\mathcal{T}_{i}}\big)   \label{eqn:imp}
\end{equation}

where $\textbf{q}_{i}$ is the probability distribution over the image categories in the dataset. $\textbf{W}_{1}, \textbf{W}_{2}$ \& $\textbf{W}_{3}$ are the training parameters of the model which are jointly-optimized
along with other model parameters. We apply the argmax operation on $\textbf{q}_{i}$ for determining the model predictions on the image label, $y^{o}_{i}$. In summary, the hierarchical graph encoder(Hg-Enc) learns the multi-spatial scale patterns in the nanoscale images. The Graph Encoder(GEnc) and clique tree encoder(CTEnc) are tailored-designed to overcome the inherent limitations of the high intra-class dissimilarity and high inter-class similarity in the nanomaterial images classification task.

\vspace{-2mm}
\section{Datasets}

We conduct experiments on the Scanning Electron Microscopy dataset\cite{aversa2018first} for nanomaterials identification. The labeled dataset contains $\approx$21,283 high-resolution images at the nanoscale. The dataset contains a wide collection of nano-objects. The images are classified into 10 categories, and they span a broad range of \textit{particles, nanowires, patterned surfaces, microelectromechanical devices}, ~\emph{etc}. The images are augmented by  geometric transformations such as shearing, and resizing to improve the quality of training data and reduce over-fitting. We randomly split the data. The test set comprises 4256 samples. The validation set contains 2128 samples, and the remaining are for the training set. The dataset is highly imbalanced. We utilize 10-Fold cross-validation for training the model to learn from the training set. In addition, we utilized several open-source material benchmark datasets to evaluate our proposed method. The details are reported in the supplementary material.

\vspace{-2mm}
\section{Experimental Setup}
\label{sec:ES} 

The model was trained in a supervised-learning approach for the classification task. The data pre-processing involves feature scaling to obtain normalized images. 
The resolution of each RGB image in our dataset is $1024\times 768\times 3$ pixels. We perform resizing of the image to obtain a relatively lower spatial resolution, $256\times 256\times 3$ pixels. We split the downscaled image into non-overlapping patches with resolution $32\times 32\times 3$ pixels. The positional and patch embedding size($\textit{d}$) is 64. The batch size is 24. The iterative message-passing steps($\text{T}$) for $\text{GEnc}$ \& $\text{CTEnc}$ operators is 6. The training number of epochs is 100. The initial learning rate is $5e^{-3}$. We reduce the learning rate by half if the evaluation metric shows no improvement on the validation set for a waiting number of 10 epochs. We run the gradient descent algorithm to minimize the cross-entropy loss between the ground-truth labels and the model predictions. We report the ensemble average of the results obtained from five computational experiments. Each computational experiment is run for a unique random seed. The experimental results reported are the average value of the different random seeds-based experimental run outputs. We utilized NVIDIA Tesla T4 GPUs for the training of GNNs built upon the PyTorch framework. Early stopping is implemented on the validation set to prevent the model from over-fitting and for model selection. We evaluate the  model performance and report the evaluation metric on the test set.

\vspace{-3mm}
\section{Experiments}

In this work, we find an answer to the following research questions.

\begin{itemize}
\item RQ1 : How does our proposed method perform in classification tasks compared to the ConvNets(CNNs), Vision Transformers(ViTs), and traditional GNNs-based algorithms$?$
\item RQ2 : How helpful are the various modules of our proposed method for the improved overall performance of the model$?$
\item RQ3 : What about the impact of positional encoding schemes$?$
\item RQ4 : How do the baselines modules of similar functionality perform compared to the modules in our framework$?$ 
\item RQ5 : What is the impact of image categories on each modules$?$
\item RQ6 : What about the quality of learned embeddings in self-supervised settings$?$
\end{itemize}

\begin{table}[htbp]
\centering
\setlength{\tabcolsep}{2pt}
\caption{Comparative study of our proposed method and the baseline algorithms. We also report the performance of the Self-Supervised(VSL) learning algorithms on Vision tasks.}
\label{tab:table1}
\vspace{-2mm}
\begin{tabular}{cc|cccccc}
\hline
\multicolumn{2}{c|}{\textbf{Algorithms}}                      &\textbf{Parameters}                & \textbf{Top-1} & \textbf{Top-2} & \textbf{Top-3} & \textbf{Top-5}  \\ \hline
\multicolumn{1}{c|}{\multirow{6}{*}{\rotatebox[origin=c]{90}{\textbf{ConvNets}}}} & AlexNet      & 5.70E+07 &   0.493                &   0.582                &   0.673                &      0.793             &                     \\
\multicolumn{1}{c|}{}                                          & DenseNet     & 2.39E+05  & 0.539             & 0.750             & 0.875             & 0.906             &             \\
\multicolumn{1}{c|}{}                                          & ResNet       & 2.72E+05  & 0.512             & 0.766             & 0.891             & 0.906             &             \\
\multicolumn{1}{c|}{}                                          & VGG          & 3.44E+07  &  0.517                 & 0.644                  & 0.717                  &                  0.779 &                     \\
\multicolumn{1}{c|}{}                                          & GoogleNet    & 2.61E+05  & 0.560             & 0.844             & 0.906             & 0.938              \\
\multicolumn{1}{c|}{}                                          & SqueezeNet   & 7.41E+05  & 0.436             & 0.469             & 0.609             & 0.656              \\ \hline
\multicolumn{1}{c|}{\multirow{6}{*}{\rotatebox[origin=c]{90}{\textbf{VSL}}}} & Barlowtwins\cite{zbontar2021barlow}  & 8.99E+06    & 0.138             & 0.250             & 0.328             & 0.453                         \\
\multicolumn{1}{c|}{}                                          & SimCLR\cite{chen2020simple}       & 8.73E+06    & 0.157             & 0.234             & 0.359             & 0.469                       \\
\multicolumn{1}{c|}{}                                          & byol\cite{grill2020bootstrap}         & 8.86E+06   & 0.130             & 0.234             & 0.281             & 0.422                         \\
\multicolumn{1}{c|}{}                                          & moco\cite{he2020momentum}         & 8.73E+06   & 0.158             & 0.188             & 0.250             & 0.438                         \\
\multicolumn{1}{c|}{}                                          & nnclr\cite{dwibedi2021little}        & 9.12E+06   & 0.144             & 0.266             & 0.313             & 0.531                         \\
\multicolumn{1}{c|}{}                                          & simsiam\cite{chen2021exploring}      & 9.01E+6   & 0.170             & 0.266             & 0.391             & 0.500                          \\ \hline
\multicolumn{1}{c|}{\multirow{24}{*}{\rotatebox[origin=c]{90}{\textbf{Vision Transformers(ViTs)}}}}        & CCT\cite{hassani2021escaping}          & 4.10E+05    & 0.600             & 0.781             & 0.875             & 0.969             &             \\
\multicolumn{1}{c|}{}                                          & CVT\cite{CVT}   & 2.56E+05    & 0.537             & 0.750             & 0.828             & 0.953               \\
\multicolumn{1}{c|}{}                                          & ConViT\cite{ConViT}       & 6.00E+05    & 0.582             & 0.734             & 0.828             & 0.938                         \\
\multicolumn{1}{c|}{}                                          & ConvVT\cite{CVT}       & 9.23E+04   & 0.291             & 0.563             & 0.734             & 0.875              \\
\multicolumn{1}{c|}{}                                          & CrossViT\cite{Crossvit}     & 8.35E+05    & 0.466             & 0.719             & 0.828             & 0.938                      \\
\multicolumn{1}{c|}{}                                          & PVTC\cite{PVT}         & 1.30E+06    & 0.567             & 0.766             & 0.813             & 0.922             &            \\
\multicolumn{1}{c|}{}                                          & SwinT\cite{SwinT}        & 2.78E+07    & 0.675             & 0.766             & 0.891             & 0.938                         \\
\multicolumn{1}{c|}{}                                          & VanillaViT\cite{dosovitskiy2020image}   & 1.79E+06    & 0.623             & 0.828             & 0.859             & 0.938              \\
\multicolumn{1}{c|}{}                                          & Visformer\cite{visformer}    & 1.21E+05   & 0.371             & 0.578             & 0.641             & 0.797                         \\ 
\multicolumn{1}{c|}{}                                          & ATS\cite{fayyaz2021ats}          & 3.26E+06     & 0.511             & 0.703             & 0.828             & 0.938                       \\
\multicolumn{1}{c|}{}                                          & CaiT\cite{CaiT}         & 3.84E+07    & 0.616             & 0.750             & 0.906             & 0.938             &            \\
\multicolumn{1}{c|}{}                                          & DeepViT\cite{Deepvit}      & 3.26E+06    & 0.512             & 0.734             & 0.875             & 0.938                         \\
\multicolumn{1}{c|}{}                                          & Dino\cite{Dino}         & 2.02E+07    & 0.047             & 0.219             & 0.391             & 0.432             &             \\
\multicolumn{1}{c|}{}                                          & Distallation\cite{Distillation} & 2.06E+06  & 0.516             & 0.719             & 0.844             & 0.938                       \\
\multicolumn{1}{c|}{}                                          & LeViT\cite{Levit}        & 1.68E+07   & 0.597             & 0.813             & 0.875             & 0.953                         \\
\multicolumn{1}{c|}{}                                          & MA\cite{MA}           & 3.87E+06    & 0.192             & 0.288             & 0.350             & 0.459             &            \\
\multicolumn{1}{c|}{}                                          & NesT\cite{Nest}         & 1.61E+07      & 0.636             & 0.828             & 0.891             & 0.953             &             \\
\multicolumn{1}{c|}{}                                          & PatchMerger\cite{PatchMerger}  & 3.26E+06    & 0.549             & 0.719             & 0.859             & 0.922                         \\
\multicolumn{1}{c|}{}                                          & PiT\cite{PiT}          & 4.48E+06     & 0.520             & 0.703             & 0.828             & 0.953             &            \\
\multicolumn{1}{c|}{}                                          & RegionViT\cite{Regionvit}    & 1.22E+07    & 0.575             & 0.797             & 0.859             & 0.922                         \\
\multicolumn{1}{c|}{}                                          & SMIM\cite{SMIM}         & 2.38E+06     & 0.163             & 0.297             & 0.453             & 0.609             &             \\
\multicolumn{1}{c|}{}                                          & T2TViT\cite{T2TViT}      & 1.03E+07    & 0.702             & 0.859             & 0.906             & 0.938                         \\
\multicolumn{1}{c|}{}                                          & ViT-SD\cite{ViT-SD}          & 4.47E+06     & 0.613             & 0.766             & 0.906             & 0.953                         \\ 
\hline
\multicolumn{1}{c|}{}                                          & \textbf{EMCNet}     & 9.70E+05     &    \textbf{0.783}               &     \textbf{0.876}              &      \textbf{0.952}             &    \textbf{0.984}               &                     \\ \hline
\end{tabular}
\end{table}

\vspace{-2mm}
\section{Results}
\subsection{RQ1: Benchmarking algorithms}
The initial experimental results on the dataset are reported by \cite{modarres2017neural}. They evaluate the well-known inception model and its variant's performance on the subset of the original dataset \cite{aversa2018first}, which contains a set of 10 categories for a total of 18,577 images. Due to the unavailability of the subset dataset publicly, we conducted experiments on the original dataset \cite{aversa2018first} which contains 12$\%$ higher samples. However, the original dataset \cite{aversa2018first} curator doesn't provide the predefined train/validation/test dataset. So, we utilize the k-fold cross-validation method to evaluate our model performance on the dataset. Table \ref{tab:table1} presents the performance of the baseline models based on ConvNets(CNNs), and Vision Transformers(ViTs) architectures in comparison with our method. We utilize the neural network architecture reported in the literature for the baseline models. For fair and rigorous comparison, we adopt identical experimental settings to generate the results of the baseline models. The evaluation metric is the conventional Top-N accuracy, where $N \in \{1, 2, 3, 5\}$. The standard deviation values are less than almost 2$\%$ of the mean value. For further comparisons with the recent advances in the GNNs. We present a reasonable comparison under identical experimental settings with the baseline GNNs techniques in Table \ref{tab:table2}. The baseline GNNs performance is evaluated on the graph representation of the nanoscale images. In comparison with the baseline models, our proposed method attains significant gains and demonstrates the best performance consistently across the Top-N accuracy classification scores. We also compare our model performance with the contrastive learning algorithms on vision and graph data for classification tasks. We report the results in Tables, \ref{tab:table1} and \ref{tab:table2}. We also evaluate our model performance in terms of precision (fraction of positive classified images are actually correct) and recall (fraction of actual positives identified correctly). We report an average precision of 0.705 and an average recall of 0.763 across the image categories.

\vspace{-3mm}
\begin{table}[htbp]
\centering
\setlength{\tabcolsep}{2pt}
\caption{Comparative study of our proposed method and the baseline algorithms. We also report the performance of the Graph Self-Supervised(GSL) learning algorithms.}
\label{tab:table2}
\vspace{-2mm}
\begin{tabular}{cc|cccccc}
\hline
\multicolumn{2}{c|}{\textbf{Algorithms}}                           &\textbf{Parameters}           & \textbf{Top-1} & \textbf{Top-2} & \textbf{Top-3} & \textbf{Top-5}  \\ \hline
\multicolumn{1}{c|}{\multirow{4}{*}{\rotatebox[origin=c]{90}{\textbf{GSL}}}} & GBT\cite{bielak2021graph}    & 7.09E+05     & 0.513             & 0.595             & 0.686             & 0.778                        \\
\multicolumn{1}{c|}{}                                          & GRACE\cite{zhu2020deep}         & 7.44E+05    & 0.581             & 0.646             & 0.711             & 0.773              \\
\multicolumn{1}{c|}{}                                          & BGRL\cite{thakoor2021bootstrapped}          & 6.92E+05    & 0.573             & 0.629             & 0.671             & 0.728                       \\
\multicolumn{1}{c|}{}                                          & InfoGraph\cite{sun2019infograph}        & 6.82E+05    & 0.560             & 0.631             & 0.694             & 0.756                 \\
\hline
\multicolumn{1}{c|}{\multirow{15}{*}{\rotatebox[origin=c]{90}{\textbf{Graph Convolution Networks}}}}        & APPNP\cite{klicpera2018predict}           & 7.35E+05    & 0.604             & 0.713             & 0.792             & 0.823                                                \\
\multicolumn{1}{c|}{}                                          & AGNN\cite{thekumparampil2018attention}  & 5.22E+05    & 0.517             & 0.733             & 0.841             & 0.943                        \\
\multicolumn{1}{c|}{}                                          & ARMA\cite{bianchi2021graph}       & 4.57E+05    & 0.553             & 0.747             & 0.848             & 0.925                         \\
\multicolumn{1}{c|}{}                                          & DNA\cite{fey2019just}        & 8.48E+05    & 0.593             & 0.677             & 0.786             & 0.891                      \\
\multicolumn{1}{c|}{}                                          & GAT\cite{velivckovic2017graph}      & 6.31E+05    & 0.507             & 0.724             & 0.807             & 0.914                         \\
\multicolumn{1}{c|}{}                                          & GGConv\cite{li2015gated}          & 8.05E+05    & 0.583             & 0.778             & 0.841             & 0.944                         \\
\multicolumn{1}{c|}{}                                          & GraphConv\cite{morris2019weisfeiler}         & 5.85E+05     & 0.623             & 0.787             & 0.875             & 0.953                         \\
\multicolumn{1}{c|}{}                                          & GCN2Conv\cite{chen}   & 6.18E+05    & 0.697             & 0.813             & 0.867             & 0.945             &            \\
\multicolumn{1}{c|}{}                                          & ChebConv\cite{defferrard2016convolutional}     & 5.00E+05    & 0.547             & 0.762             & 0.834 & 0.896                                                \\ 
\multicolumn{1}{c|}{}                                          & GraphConv\cite{morris2019weisfeiler}           & 6.79E+05    & 0.533             & 0.727             & 0.847             & 0.961                         \\
\multicolumn{1}{c|}{}                                          & GraphUNet\cite{gao2019graph}        & 9.57E+05    & 0.622             & 0.738             & 0.866             & 0.912                         \\
\multicolumn{1}{c|}{}                                          & MPNN\cite{gilmer2017neural}       & 5.22E+05    & 0.643             & 0.792             & 0.873             & 0.959                         \\
\multicolumn{1}{c|}{}                                          & RGGConv\cite{bresson2017residual}          & 6.58E+05    & 0.633             & 0.727             & 0.886             & 0.928                        \\
\multicolumn{1}{c|}{}                                          & SuperGAT\cite{kim2022find}  & 5.54E+05    & 0.561             & 0.676             & 0.863             & 0.935                         \\
\multicolumn{1}{c|}{}                                          & TAGConv\cite{du2017topology}         & 5.74E+05    & 0.614             & 0.739             & 0.803             & 0.946                         \\
\hline
\multicolumn{1}{c|}{}                                          & \textbf{EMCNet}      & 9.70E+05    &    \textbf{0.783}               &     \textbf{0.876}              &      \textbf{0.952}             &    \textbf{0.984}               &                     \\ \hline
\end{tabular}
\end{table}

\subsection{RQ2: Study of Modules}
Our proposed method comprises of $\textbf{GEnc}$, $\textbf{HGEnc}$, and $\textbf{CTEnc}$ modules. We perform experiments on the SEM dataset by gradually removing each module in a controlled setting to examine the efficacy of the key modules that be responsible for the improved overall performance of the model. We refer to the $\textbf{EMCNet}$ model in the absence of different operators as follows:

\begin{itemize}
\item $\textbf{w/o} \hspace{1mm} \textbf{GEnc}$: $\textbf{EMCNet}$ model without the $\textbf{GEnc}$ operator.
\item $\textbf{w/o} \hspace{1mm} \textbf{HGEnc}$: $\textbf{EMCNet}$ model without the $\textbf{HGEnc}$ operator.
\item $\textbf{w/o} \hspace{1mm} \textbf{CTEnc}$: $\textbf{EMCNet}$ model without the $\textbf{CTEnc}$ operator. 
\end{itemize}

We report in Table ~\ref{table:main_3} the results obtained on the test dataset in terms of Top-1 classification accuracy. The results demonstrate the effectiveness of the operators with negligible add-on computational complexity. To be specific, the $\textbf{HGEnc}$ operator is of advantage in learning the long-range dependencies and complex hierarchical representations of the graph in comparison to the flat message-passing schemes in GNNs. Similarly, $\textbf{CTEnc}$ operator learns the complex structural characteristics in the graphs through the tree-structured scaffolds representations. The $\textbf{GEnc}$ operator helps in better learning of the rich local-graph neighborhood information.

\vspace{-2mm}
\begin{table}[htbp]
\centering
\setlength{\tabcolsep}{2pt}
\caption{The table reports the ablation studies to characterize the effect of each module on the overall model performance.}
\vspace{-3mm}
\label{table:main_3}
\begin{adjustbox}{width=1\columnwidth,center}
{%
\begin{tabular}{@{}c|c|c|c|c@{}}
\hline
Methods & $\textbf{EMCNet}$ & $\textbf{w/o} \hspace{1mm} \textbf{GEnc}$  & $\textbf{w/o} \hspace{1mm} \textbf{HGEnc}$  & $\textbf{w/o}\hspace{1mm}\textbf{CTEnc}$  \\
\hline
Accuracy & \textbf{0.783} $\pm$ 0.012 & \textbf{0.717} $\pm$ 0.014 &  \textbf{0.594} $\pm$ 0.015 &  \textbf{0.664} $\pm$ 0.019  \\
\hline
\end{tabular}
}
\end{adjustbox}

\end{table}

\vspace{-3mm}
\subsection{RQ3: Study on Tokenization of Images}

We conduct experiments to investigate the impact of positional embeddings that contribute to the better performance of the model. We refer to the $\textbf{EMCNet}$ model without incorporating the positional embeddings as follows: $\textbf{w/o} \hspace{1mm} \textbf{PosEmb}$: $\textbf{EMCNet}$ model. We report in Table ~\ref{table:main_4} the results obtained on the test dataset in terms of Top-1 accuracy. The impact is significant on the model performance.

\vspace{-2mm}
\begin{table}[htbp]
\centering
\caption{The ablation studies to determine the impact of positional embeddings on the model performance.}
\label{table:main_4}
\vspace{-3mm}
\begin{adjustbox}{width=0.65\columnwidth,center}
{%
\begin{tabular}{@{}c|c|cc@{}}
\hline
Methods & $\textbf{EMCNet}$ & $\textbf{w/o} \hspace{1mm} \textbf{PosEmb}$  &   \\
\hline
Accuracy & \textbf{0.783} $\pm$ 0.012 &  \textbf{0.515} $\pm$ 0.012  &   \\
\hline
\end{tabular}
}
\end{adjustbox}
\end{table}

\vspace{-2mm}
We further investigate the model performance by modeling the positional encoding scheme(PEs) by the Laplacian PEs\cite{dwivedi2020generalization}, Self-attention RPEs,\cite{shaw2018self}, Signed RPEs\cite{huang2020improve}, and Sinusoidal PEs\cite{vaswani2017attention} compared to our approach of learning the optimal position embeddings through training parameters of the model. We refer to the $\textbf{EMCNet}$ model by incorporating the following PEs techniques as follows:

\begin{itemize}
\item \hspace{-1.0mm}$\textbf{w/} \hspace{1mm} \textbf{LPEs}$: $\textbf{EMCNet}$ model with \textbf{Laplacian PEs}.
\item \hspace{-1mm}$\textbf{w/} \hspace{1mm} \textbf{SPEs}$: $\textbf{EMCNet}$ model with \textbf{Sinusoidal PEs}.
\item \hspace{-1.0mm}$\textbf{w/} \hspace{1mm} \textbf{SARPEs}$: $\textbf{EMCNet}$ model with \textbf{Self-attention RPEs}.
\item \hspace{-1mm}$\textbf{w/} \hspace{1mm} \textbf{APEs}$: $\textbf{EMCNet}$ model with \textbf{Signed RPEs}.
\end{itemize}

We report in Table ~\ref{table:main_5} the results on the test dataset in terms of Top-1 classification accuracy. The impact of various PEs techniques is marginal on the model predictive performance in comparison with our method.

\vspace{-2mm}
\begin{table}[htbp]
\setlength{\tabcolsep}{2pt}
\caption{The table reports the ablation studies on positional-encoding schemes.}
\label{table:main_5}
\vspace{-3mm}
\begin{adjustbox}{width=1.05\columnwidth,center}
{%
\hspace{-8mm}\begin{tabular}{@{}c|c|c|c|c|cc@{}}
\hline
Methods & $\textbf{EMCNet}$ & $\textbf{w/} \hspace{1mm} \textbf{LPEs}$  & $\textbf{w/} \hspace{1mm} \textbf{SPEs}$ & $\textbf{w/} \hspace{1mm} \textbf{SARPEs}$  & $\textbf{w/} \hspace{1mm} \textbf{SRPEs}$ \\
\hline
Accuracy & \textbf{0.783} $\pm$ 0.012 &  \textbf{0.772} $\pm$ 0.008  &  \textbf{0.767} $\pm$ 0.010  &  \textbf{0.790} $\pm$ 0.003  &  \textbf{0.761} $\pm$ 0.007 \\
\hline
\end{tabular}
}
\end{adjustbox}

\end{table}

\vspace{-4mm}
\subsection{RQ4: Study of GEnc module}

We investigate the GEnc module efficacy in comparison with the popular algorithms of similar functionality. The GEnc module consists of the graph convolution operator to perform convolutional operations on graphs and the global average pooling operator to determine the graph-level representation. We utilize well-known methods as baseline operators to perform convolution on the graphs. The list includes,  GAT(\cite{velivckovic2017graph}), APPNP(\cite{klicpera2018predict}), DNA(\cite{fey2019just}), and GCN2(\cite{chen2020}). For the global graph-pooling, the list includes GraphMultisetTransformer(GMT, \cite{baek2021accurate}), GlobalAttention(GA, \cite{li2015gated}), Set2Set(\cite{vinyals2015order}), and Global Summation Pooling(GSM). We refer to the $\textbf{EMCNet}$ model with the baseline graph convolution operators are as follows:

\begin{itemize}
\item $\textbf{w/} \hspace{1mm} \textbf{GAT}$:  The $\textbf{EMCNet}$ model with $\textbf{GAT}$ operator.
\item $\textbf{w/} \hspace{1mm} \textbf{APPNP}$:  The $\textbf{EMCNet}$ model with $\textbf{APPNP}$ operator.
\item $\textbf{w/} \hspace{1mm} \textbf{DNA}$:  The $\textbf{EMCNet}$ model with  $\textbf{DNA}$ operator.
\item $\textbf{w/} \hspace{1mm} \textbf{GCN2}$:  The $\textbf{EMCNet}$ model with  $\textbf{GCN2}$ operator.
\end{itemize}

The Top-1 accuracy is reported in Table ~\ref{table:main_6}. The results support our graph convolution operator based approach to overcome the shallow learning mechanisms of the baseline operators. 

\vspace{-2mm}
\begin{table}[htbp]
\centering
\setlength{\tabcolsep}{2pt}
\caption{The table reports the comparative study of the graph convolution baseline operators on the model performance.}
\label{table:main_6}
\vspace{-3mm}
\begin{adjustbox}{width=1.05\columnwidth,center}
{%
\hspace{-10mm}\begin{tabular}{@{}c|c|c|c|c|c@{}}
\hline
Methods & $\textbf{EMCNet}$ & $\textbf{w/} \hspace{1mm} \textbf{GAT}$  & $\textbf{w/} \hspace{1mm} \textbf{APPNP}$ & $\textbf{w/} \hspace{1mm} \textbf{DNA}$  & $\textbf{w/} \hspace{1mm} \textbf{GCN2}$ \\
\hline
Accuracy &  \textbf{0.783} $\pm$ 0.012 &  \textbf{0.747} $\pm$ 0.012  &  \textbf{0.759} $\pm$ 0.010  &  \textbf{0.767} $\pm$ 0.009  &  \textbf{0.749} $\pm$ 0.013 \\
\hline
\end{tabular}
}
\end{adjustbox}

\end{table}

\vspace{-2mm}
We refer to the $\textbf{EMCNet}$ model with the baseline operators for modeling the global read-out function in the $\textbf{GEnc}$ module as:

\begin{itemize}
\item $\textbf{w/} \hspace{1mm} \textbf{GMT}$: The $\textbf{EMCNet}$ model with $\textbf{GMT}$ operator in $\textbf{GEnc}$.
\item $\textbf{w/} \hspace{1mm} \textbf{GA}$: The $\textbf{EMCNet}$ model with $\textbf{GA}$ operator in $\textbf{GEnc}$.
\item $\textbf{w/} \hspace{1mm} \textbf{Set2Set}$: The $\textbf{EMCNet}$ model with $\textbf{Set2Set}$ in $\textbf{GEnc}$.
\item $\textbf{w/} \hspace{1mm} \textbf{GSM}$: The $\textbf{EMCNet}$ model with $\textbf{GSM}$ operator in $\textbf{GEnc}$.
\end{itemize}

\vspace{0mm}
We report the results obtained on the test dataset in terms of Top-1 classification accuracy. The results reported in Table ~\ref{table:main_7} demonstrate no significant improvements in the model performance in comparison to our method with the global average pooling operator. Overall, our graph encoder proves to be effective by learning to compute optimal graph-level representations.

\vspace{-2mm}
\begin{table}[htbp]
\centering
\setlength{\tabcolsep}{2pt}
\caption{The table reports the comparative study of the baseline global pooling operators on the model performance}
\label{table:main_7}
\vspace{-3mm}
\begin{adjustbox}{width=1.05\columnwidth,center}
{%
\hspace{-10mm}\begin{tabular}{@{}c|c|c|c|c|c@{}}
\hline
Methods & $\textbf{EMCNet}$ & $\textbf{w/} \hspace{1mm} \textbf{GMT}$  & $\textbf{w/} \hspace{1mm} \textbf{GA}$ & $\textbf{w/} \hspace{1mm} \textbf{Set2Set}$  & $\textbf{w/} \hspace{1mm} \textbf{GSM}$ \\
\hline
Accuracy & \textbf{0.783} $\pm$ 0.012 &  \textbf{0.797} $\pm$ 0.011  &  \textbf{0.791} $\pm$ 0.007  &  \textbf{0.786} $\pm$ 0.008  &  \textbf{0.772} $\pm$ 0.013 \\
\hline
\end{tabular}
}
\end{adjustbox}
\end{table}

\vspace{-4mm}
\subsection{RQ4: Study of $\textbf{HGEnc}$ module}

We probe the HGEnc module effectiveness in comparison with the well-known algorithms of identical functionality. The $\textbf{HGEnc}$ module operates in two phases. The first phase performs the graph coarsening and the high-order message-passing schemes to compute the hierarchical representations of the nodes in the coarser graph. The successive phase performs the global average pooling of the node representations to compute the graph-level representation. We utilize popular methods as the baseline operators for modeling the hierarchical graph coarsening and message-passing schemes. The list includes,  MemPool(\cite{velivckovic2017graph}), ASAPool(\cite{klicpera2018predict}), SAGPool(\cite{fey2019just}), and TopKPool(\cite{gao2019graph}). We refer to the $\textbf{EMCNet}$ algorithm with the baseline hierarchical graph convolution operators are as follows:

\begin{itemize}
\item $\textbf{w/} \hspace{1mm} \textbf{MemPool}$:  $\textbf{EMCNet}$ model with $\textbf{MemPool}$ operator.
\item $\textbf{w/} \hspace{1mm} \textbf{ASAPool}$:  $\textbf{EMCNet}$ model with $\textbf{ASAPool}$ operator.
\item $\textbf{w/} \hspace{1mm} \textbf{SAGPool}$:  $\textbf{EMCNet}$ model with  $\textbf{SAGPool}$ operator.
\item $\textbf{w/} \hspace{1mm} \textbf{TopKPool}$:  $\textbf{EMCNet}$ model with  $\textbf{TopKPool}$ operator.
\end{itemize}

\vspace{-3mm}
\begin{table}[htbp]
\centering
\setlength{\tabcolsep}{2pt}
\caption{The table reports the study of baseline hierarchical graph convolution operators on the model performance.}
\label{table:main_8}
\vspace{-3mm}
\begin{adjustbox}{width=1.25\columnwidth,center}
{%
\hspace{10mm}\begin{tabular}{@{}c|c|c|c|c|c@{}}
\hline
Methods & $\textbf{EMCNet}$ & $\textbf{w/} \hspace{1mm} \textbf{MemPool}$  & $\textbf{w/} \hspace{1mm} \textbf{ASAPool}$ & $\textbf{w/} \hspace{1mm} \textbf{SAGPool}$  & $\textbf{w/} \hspace{1mm} \textbf{TopKPool}$ \\
\hline
Accuracy & \textbf{0.783} $\pm$ 0.012 &  \textbf{0.713} $\pm$ 0.014  &  \textbf{0.725} $\pm$ 0.011  &  \textbf{0.709} $\pm$ 0.008  &  \textbf{0.725} $\pm$ 0.013 \\
\hline
\end{tabular}
}
\end{adjustbox}

\end{table}

\vspace{-3mm}
The Top-1 classification accuracy is reported in Table ~\ref{table:main_8} on the test dataset. The results demonstrate the efficacy of $\textbf{HGEnc}$ module to model the complex local substructures to learn the hierarchical representations of the graph.  We refer to the $\textbf{EMCNet}$ model with the following baseline operators to perform the global average pooling in the $\textbf{HGEnc}$ module as:

\begin{itemize}
\item $\textbf{w/} \hspace{1mm} \textbf{GMT}$: The $\textbf{EMCNet}$ model with $\textbf{GMT}$ operator in $\textbf{HGEnc}$.
\item $\textbf{w/} \hspace{1mm} \textbf{GA}$: The $\textbf{EMCNet}$ model with $\textbf{GA}$ operator in $\textbf{HGEnc}$.
\item $\textbf{w/} \hspace{1mm} \textbf{Set2Set}$: The $\textbf{EMCNet}$ model with $\textbf{Set2Set}$ in $\textbf{HGEnc}$.
\item $\textbf{w/} \hspace{1mm} \textbf{GSM}$: The $\textbf{EMCNet}$ model with $\textbf{GSM}$ operator in $\textbf{HGEnc}$.
\end{itemize}
The results are reported on the test dataset in Table ~\ref{table:main_9} in terms of Top-1 classification accuracy. It shows no significant improvement or deterioration in the model performance in comparison to our method with the global average pooling operator.

\vspace{-1mm}
\begin{table}[htbp]
\centering
\setlength{\tabcolsep}{2pt}
\caption{The table reports the comparative study of the graph read-out baseline operators on the model performance.}
\label{table:main_9}
\vspace{-3mm}
\begin{adjustbox}{width=1.25\columnwidth,center}
{%
\hspace{10mm}\begin{tabular}{@{}c|c|c|c|c|c@{}}
\hline
Methods & $\textbf{EMCNet}$ & $\textbf{w/} \hspace{1mm} \textbf{GMT}$  & $\textbf{w/} \hspace{1mm} \textbf{GA}$ & $\textbf{w/} \hspace{1mm} \textbf{Set2Set}$  & $\textbf{w/} \hspace{1mm} \textbf{GSM}$ \\
\hline
Accuracy & \textbf{0.783} $\pm$ 0.012 &  \textbf{0.779} $\pm$ 0.006  &  \textbf{0.789} $\pm$ 0.003  &  \textbf{0.775} $\pm$ 0.011  &  \textbf{0.790} $\pm$ 0.007 \\
\hline
\end{tabular}
}
\end{adjustbox}
\end{table}

\vspace{-5mm}
\subsection{RQ5: Study of image categories impact on the modules}
Each image category in the SEM dataset consists of a mixture of easy(visually similar images) and hard samples(visually diverse), characterized by the complexity of multi-scale spatial features, degree of detail, and information density. The hard samples greatly influence learning the parameters of the modules. Similarly, each module learns distinct and dominant patterns from easy samples and generalization ability from the hard samples of each image category. Our proposed framework generalizes well despite the complexity of patterns across the wide spectrum of image categories. We study the impact of the GEnc, HGEnc, and CTEnc modules in isolation with the absence of other modules on the classification task of each image category. We perform additional experiments on the SEM dataset by realizing our proposed method with the module under investigation to support the rationale of each module on the classification task. We refer to the EMCNet model realized with the different operators as follows:

\begin{itemize}
\item $\textbf{w/} \hspace{1mm} \textbf{GEnc}$: $\textbf{EMCNet}$ model with the $\textbf{GEnc}$ operator.
\item $\textbf{w/} \hspace{1mm} \textbf{HGEnc}$: $\textbf{EMCNet}$ model with the $\textbf{HGEnc}$ operator.
\item $\textbf{w/} \hspace{1mm} \textbf{CTEnc}$: $\textbf{EMCNet}$ model with the $\textbf{CTEnc}$ operator.
\end{itemize}

We utilize the identical experimental settings, refer to section ~\ref{sec:ES} for performing this study. We report the results in Table \ref{tab:main_extra}, obtained on each image category of the test dataset in terms of Top-1 classification accuracy. We utilize 10-fold cross-validation to evaluate the model performance. The GEnc and CTEnc modules effectively capture the rich local information spread across each image for the image categories such as \textit{biological, tips, films coated surface, and powder}. The long-range and hierarchical information inherent in image categories such as \textit{fibres, porous sponge, patterned surface, nanowires, particles, MEMS devices, and powder} are effectively modeled by the HGEnc module.

\vspace{-3mm}
\begin{table}[ht]
\centering
\caption{The table reports the efficacy of each module on learning different image categories. The underline score represents the best performance among the modules.}
\label{tab:main_extra}
\vspace{-3mm}
\begin{tabular}{@{}c|ccc|c@{}}
\toprule
\multirow{2}{*}{\textbf{Category}}   & \multicolumn{3}{c|}{\textbf{Modules}}                                   & \multirow{2}{*}{\textbf{EMCNet}} \\ \cmidrule(lr){2-4}
                            & \multicolumn{1}{c|}{\textbf{GEnc}} & \multicolumn{1}{c|}{\textbf{HGEnc}} & \textbf{CTEnc} &                         \\ \midrule
Biological                  & \multicolumn{1}{c|}{\underline{0.658}}     & \multicolumn{1}{c|}{0.525}      & 0.628      & \textbf{0.727}                        \\
Tips                        & \multicolumn{1}{c|}{\underline{0.712}}     & \multicolumn{1}{c|}{0.678}      & 0.708      & \textbf{0.811}                       \\
Fibres                      & \multicolumn{1}{c|}{0.485}     & \multicolumn{1}{c|}{\underline{0.657}}      & 0.523       & \textbf{0.783}                         \\
Porous Sponge               & \multicolumn{1}{c|}{0.567}     & \multicolumn{1}{c|}{\underline{0.665}}      & 0.595      & \textbf{0.722}                        \\
Films Coated Surface        & \multicolumn{1}{c|}{0.623}     & \multicolumn{1}{c|}{0.537}      & \underline{0.679}      & \textbf{0.725}                         \\
Patterned surface           & \multicolumn{1}{c|}{0.576}     & \multicolumn{1}{c|}{\underline{0.712}}      & 0.625      & \textbf{0.771}                        \\
Nanowires                   & \multicolumn{1}{c|}{0.624}     & \multicolumn{1}{c|}{\underline{0.749}}      & 0.671      & \textbf{0.789}                        \\
Particles                   & \multicolumn{1}{c|}{0.563}     & \multicolumn{1}{c|}{\underline{0.673}}      & 0.654      & \textbf{0.729}                        \\
MEMS devices                & \multicolumn{1}{c|}{0.625}     & \multicolumn{1}{c|}{\underline{0.748}}      & 0.658      & \textbf{0.812}                        \\
Powder                      & \multicolumn{1}{c|}{\underline{0.672}}     & \multicolumn{1}{c|}{0.613}      & 0.648      & \textbf{0.771}                        \\ \bottomrule
\end{tabular}

\end{table}

\vspace{-5mm}
\subsection{RQ6: Self-Supervised Learning}
The graph contrastive learning(GCL) algorithms construct numerous arbitrary-sized graph views through stochastic data augmentation techniques. GCL is a self-supervised learning algorithm that learns a discriminative model by contrasting multiple positive-paired augmented views of the same graph, as opposed to the independently sampled negative-paired views of different graphs in the embedded space. We utilize GCL as a pre-training model to learn the node- and graph-level unsupervised representations through mutual information maximization, while optimally preserving the structural and attributional characteristics of the graph. We model the graph encoder of the contrastive techniques with our proposed, $\textbf{GEnc}$ or  $\textbf{HEnc}$ operators. We leverage a broad range of ML techniques such as Support Vector Machines, Random Forest, and the Gradient Boosted Trees for the downstream task of modeling the target label of the graph data as a function of the generated unsupervised representations. We evaluate and report the results on the test dataset in terms of Top-1 classification accuracy. The experimental results are reported in Table ~\ref{table:main_10} and Table ~\ref{table:main_11}, where we additionally report EMCNet model performance in comparison with Graph contrastive techniques. The classification scores show the effectiveness of the latent representations computed by the contrastive techniques through our design variants-based approach.

\vspace{-3mm}
\begin{table}[ht]
\centering
\caption{The table reports the performance of the contrastive techniques. The graph encoder of the contrastive techniques is modeled by $\textbf{GEnc}$ operator.}
\label{table:main_10}
\vspace{-3mm}
\begin{adjustbox}{width=1.025\columnwidth,center}
{%
\begin{tabular}{c|c|c|c|c}
\hline
Methods                                                     & $\textbf{SVM}$   & $\textbf{RF}$    & $\textbf{XGBoost}$ & $\textbf{EMCNet}$                                            \\ \hline
BGRL, \cite{thakoor2021bootstrapped} & 0.674 $\pm$ 0.09 & 0.721 $\pm$ 0.09 & 0.744 $\pm$ 0.08   & \multirow{4}{*}{\textbf{0.783} $\pm$ 0.012} \\
GBT, \cite{bielak2021graph}          & 0.693 $\pm$ 0.03 & 0.737 $\pm$ 0.05 & 0.756 $\pm$ 0.12   &                                                              \\
GRACE, \cite{zhu2020deep}            & 0.713 $\pm$ 0.05 & 0.742 $\pm$ 0.06 & 0.773 $\pm$ 0.04   &                                                              \\
InfoGraph, \cite{sun2019infograph}   & 0.723 $\pm$ 0.03 & 0.751 $\pm$ 0.11 & 0.764 $\pm$ 0.06   &                                                              \\ \hline
\end{tabular}
}
\end{adjustbox}

\end{table}

\vspace{-6mm}
\begin{table}[ht]
\centering
\caption{The table reports the performance of the contrastive techniques. The graph encoder of the contrastive techniques is modeled by $\textbf{HEnc}$ operator.}
\label{table:main_11}
\vspace{-3mm}
\begin{adjustbox}{width=1.025\columnwidth,center}
{%
\begin{tabular}{c|c|c|c|c}
\hline
Methods                                               & $\textbf{SVM}$   & $\textbf{RF}$    & $\textbf{XGBoost}$ & $\textbf{EMCNet}$                                            \\ \hline
BGRL, \cite{thakoor2021bootstrapped} & 0.693 $\pm$ 0.07 & 0.742 $\pm$ 0.09 & 0.769 $\pm$ 0.013  & \multirow{4}{*}{\textbf{0.783} $\pm$ 0.012} \\
GBT, \cite{bielak2021graph}          & 0.701 $\pm$ 0.05 & 0.758 $\pm$ 0.03 & 0.761 $\pm$ 0.06   &                                                              \\
GRACE, \cite{zhu2020deep}            & 0.717 $\pm$ 0.09 & 0.762 $\pm$ 0.09 & 0.771 $\pm$ 0.09   &                                                              \\
InfoGraph, \cite{sun2019infograph}   & 0.724 $\pm$ 0.08 & 0.752 $\pm$ 0.04 & 0.775  $\pm$ 0.06  &                                                              \\ \hline
\end{tabular}
}
\end{adjustbox}
\end{table}

\vspace{-5mm}
\subsection{Additional Experiments}
Here, we perform an additional set of experiments by resizing the input images of resolution $1024\times 768\times 3$ pixels to obtain a relatively higher spatial resolution of $512\times 512\times 3$ pixels. Here, we consider two experimental settings as follows:

\begin{itemize}
\item In the first setting(FS), we split the image into the patches of the resolution, $64\times 64\times 3$ pixels. We project the flattened patch representations into the embedding space to obtain a size of $144\times 64$.
\item In the second setting(SS), we split the image into the patches of the resolution, $32\times 32\times 3$ pixels. We project the flattened patch representations into the embedding space to obtain a size of $256\times 64$.
\end{itemize}

We refer to the $\textbf{EMCNet}$ model with the aforementioned experimental settings as follows,

\begin{itemize}
\item $\textbf{w/} \hspace{1mm} \textbf{FS}$: $\textbf{EMCNet}$ model trained with the first settings.
\item $\textbf{w/} \hspace{1mm} \textbf{SS}$: $\textbf{EMCNet}$ model trained with the second settings.
\end{itemize}

\vspace{-4mm}
\begin{table}[ht]
\centering
\caption{The table reports the effect of the patch size and the number of patches on the model performance.}
\label{table:AExp}
\vspace{-3mm}
\begin{adjustbox}{width=0.85\columnwidth,center}
{%
\begin{tabular}{@{}c|c|c|cc@{}}
\hline
Methods & $\textbf{EMCNet}$ & $\textbf{w/} \hspace{1mm} \textbf{FS}$  & $\textbf{w/} \hspace{1mm} \textbf{SS}$   \\
\hline
Accuracy & \textbf{0.783} $\pm$ 0.012 &  \textbf{0.846} $\pm$ 0.009  &  \textbf{0.867} $\pm$ 0.005   \\
\hline
\end{tabular}
}
\end{adjustbox}
\end{table}

\vspace{-6mm}
\begin{table}[ht]
\centering
\caption{Performance comparison of our clustering approach based EMCNet and the baseline EMCNet models.}
\label{table:TwoExp}
\vspace{-3mm}
\begin{adjustbox}{width=0.78\columnwidth,center}
{%
\begin{tabular}{@{}c|c|cc@{}}
\hline
Methods & $\textbf{EMCNet}$ & $\textbf{w/} \hspace{1mm} \textbf{EMCNet : Clusters}$    \\
\hline
Accuracy & \textbf{0.783} $\pm$ 0.012 &  \textbf{0.921} $\pm$ 0.0013   \\
\hline
\end{tabular}
}
\end{adjustbox}

\end{table}

\vspace{-3mm}
The results reported in Table ~\ref{table:AExp} show significant improvements in the model performance. To overcome the limitations of learning from images having a high degree of similarity corresponding to different image categories. We perform the K-Means clustering on the SEM dataset to assign cluster labels to the images based on similarity. We partition the images into k-fixed apriori clusters in which each image belongs to a specific cluster. In this work, the optimal setting for $\text{k}$ is $\textbf{2}$. We will train a single EMCNet model to learn from the images corresponding to each cluster in the supervised learning approach.  We utilize an identical train/validation/test split of 70$\%$/20$\%$/10$\%$ for images assigned to each cluster. We report the results in Table ~\ref{table:TwoExp} denoted by $\textbf{w/} \hspace{1mm} \textbf{EMCNet : Clusters}$. The results demonstrate exceptional performance in the Top-1 classification accuracy in comparison with the baseline EMCNet model. 

\vspace{-3mm}
\section{Conclusion and Future Works}

We conduct the first in-depth study of the graph-neural networks for solving the challenging electron micrographs classification tasks. We propose a joint optimization framework to determine the expressive graph-level representations by effectively summarizing the complex hierarchical visual feature maps from electron micrographs. The experimental results support our framework efficacy to achieve better performance in comparison to the state-of-art methods. In the future, we would like to extend our work on other electron micrograph datasets like REM, SEM, STEM etc., and also develop graph neural network-based models for object detection, segmentation for nanomaterial microstructures.

\vspace{-2mm}
\bibliographystyle{ACM-Reference-Format}
\bibliography{sample-base}

\section{Supplementary Material}

\subsection{SEM dataset}
\begin{figure}[htbp]
     \captionsetup[subfigure]{aboveskip=-0.15pt,belowskip=2pt}
     \centering
     \begin{subfigure}[b]{0.18\textwidth}
         \centering
         \includegraphics[width=\textwidth]{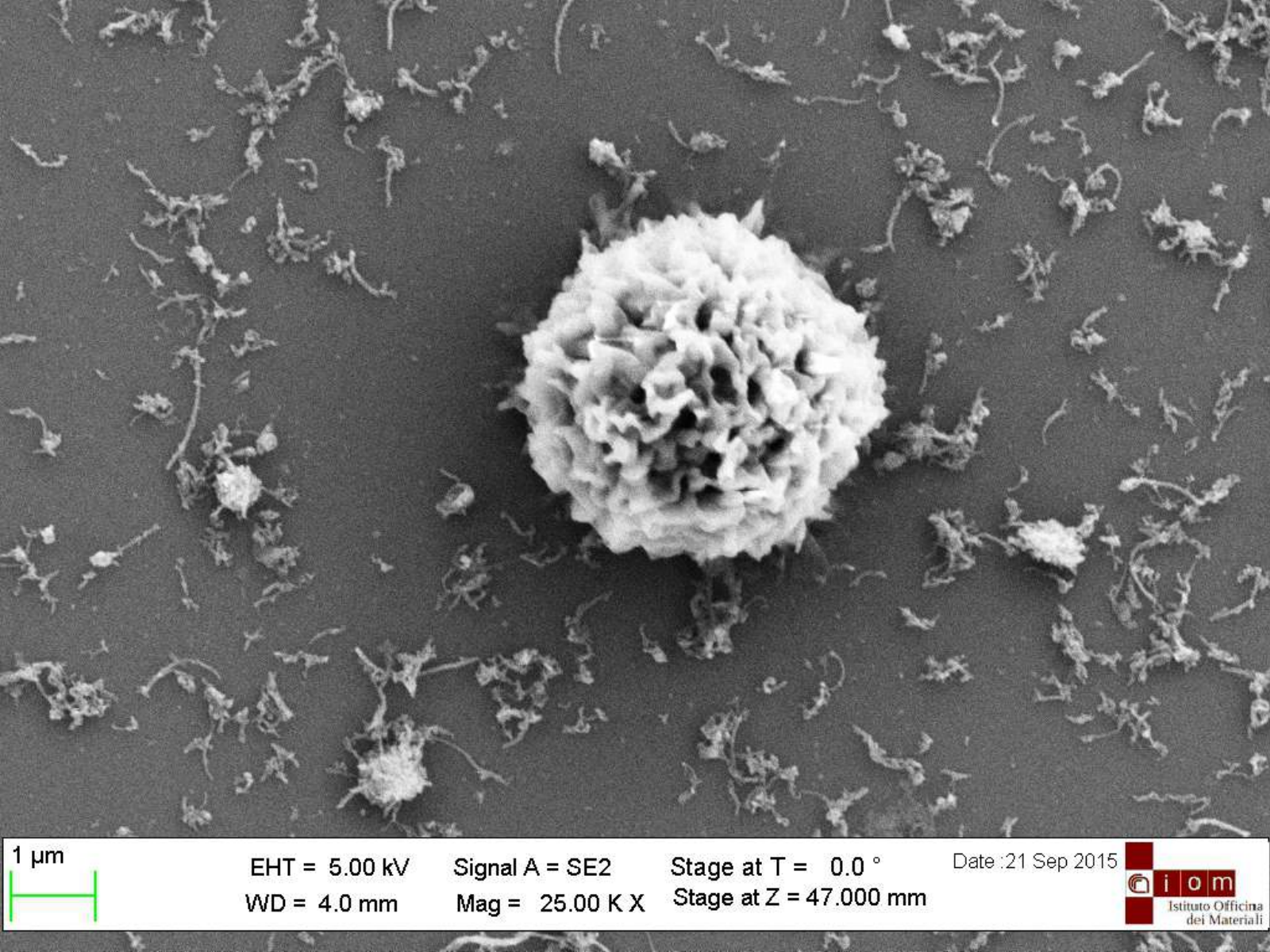}
         \caption{Biological}
         \label{fig:Biological}
     \end{subfigure}
     \hspace{1em}
     \begin{subfigure}[b]{0.18\textwidth}
         \centering
         \includegraphics[width=\textwidth]{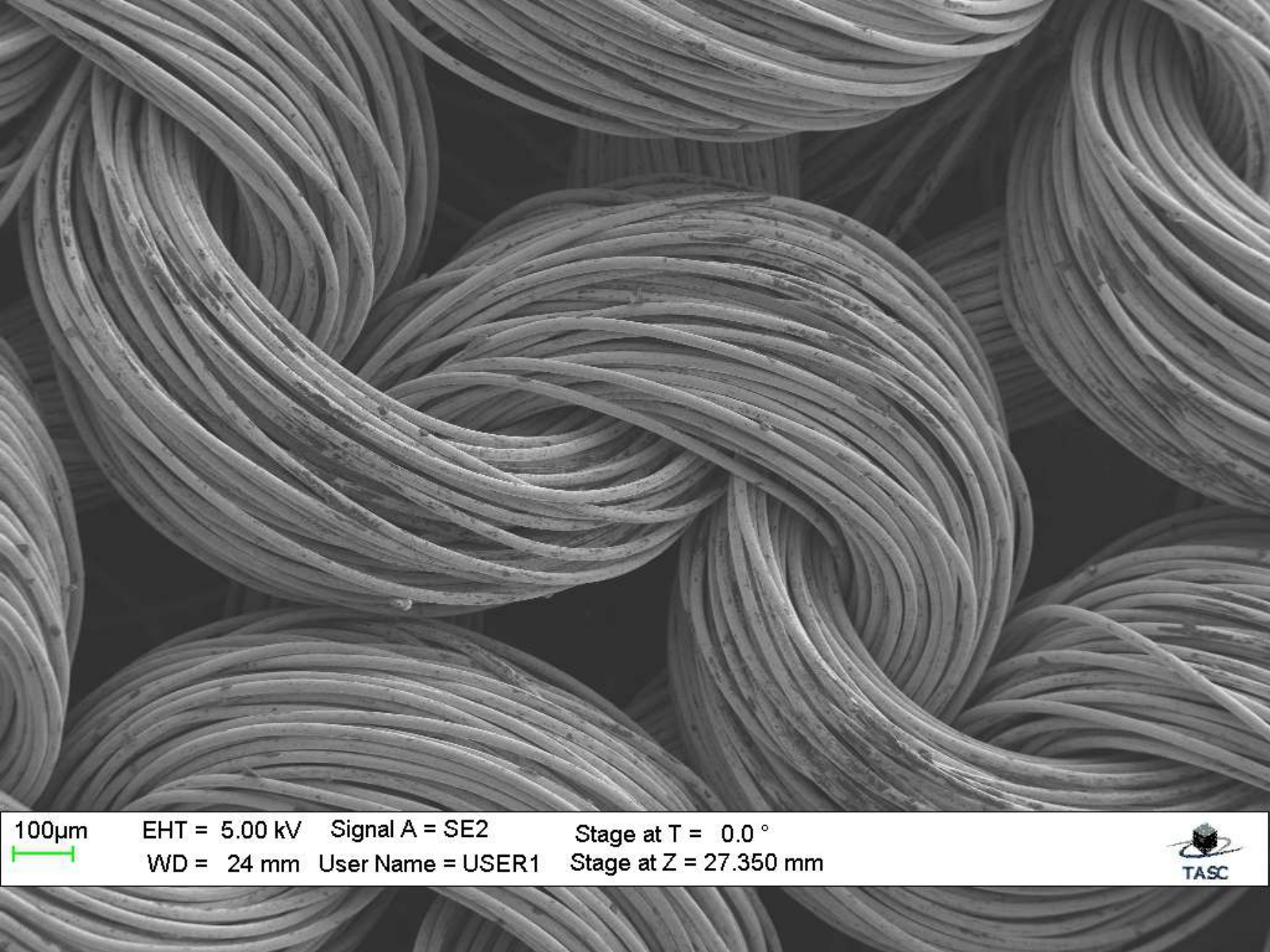}
         \caption{Fibre}
         \label{fig:fibre}
     \end{subfigure}
     \hspace{1em}
     \begin{subfigure}[b]{0.18\textwidth}
         \centering
         \includegraphics[width=\textwidth]{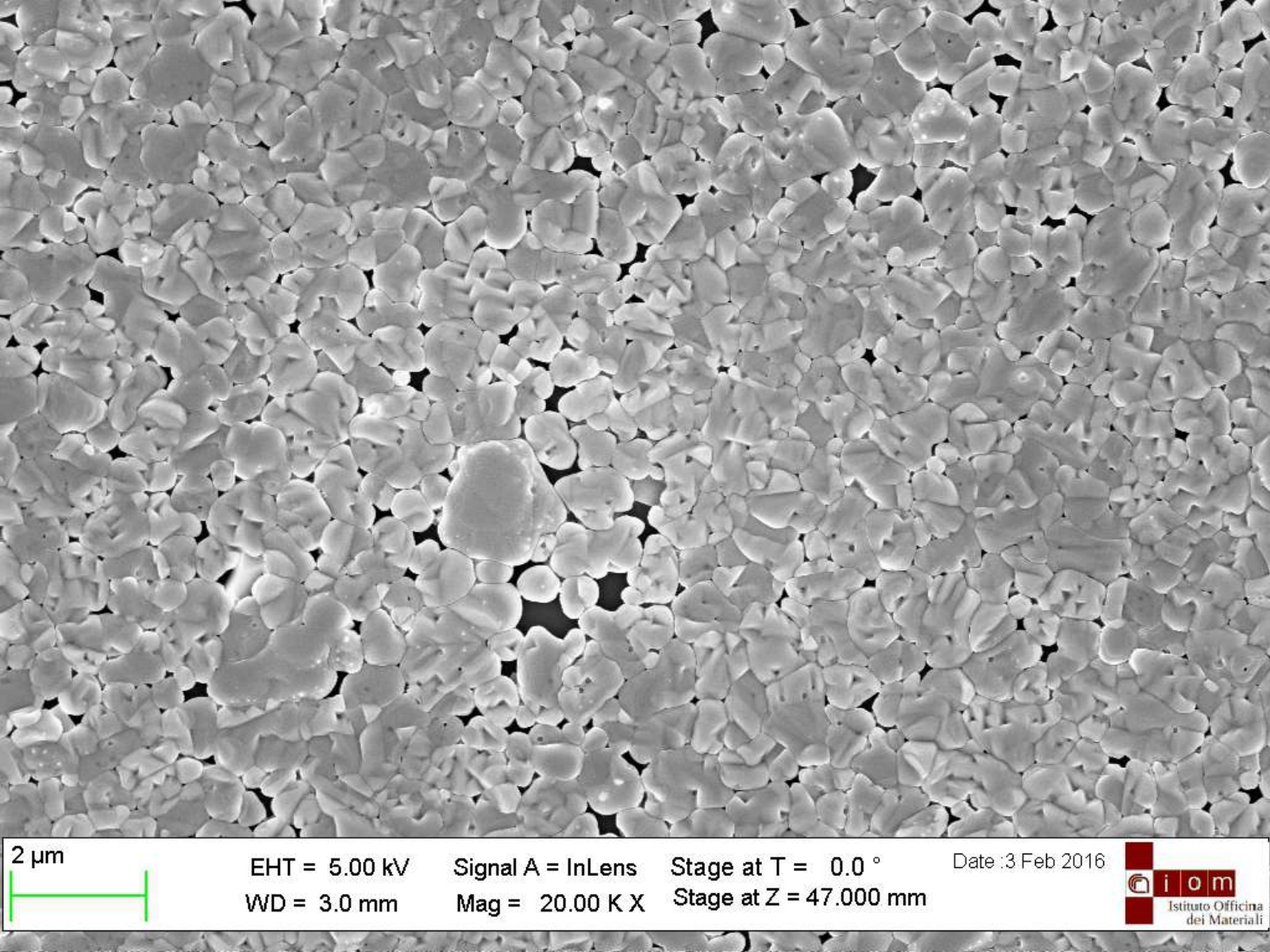}
         \caption{Films}
         \label{fig:Films}
     \end{subfigure}
     \hspace{1em}
     \begin{subfigure}[b]{0.18\textwidth}
         \centering
         \includegraphics[width=\textwidth]{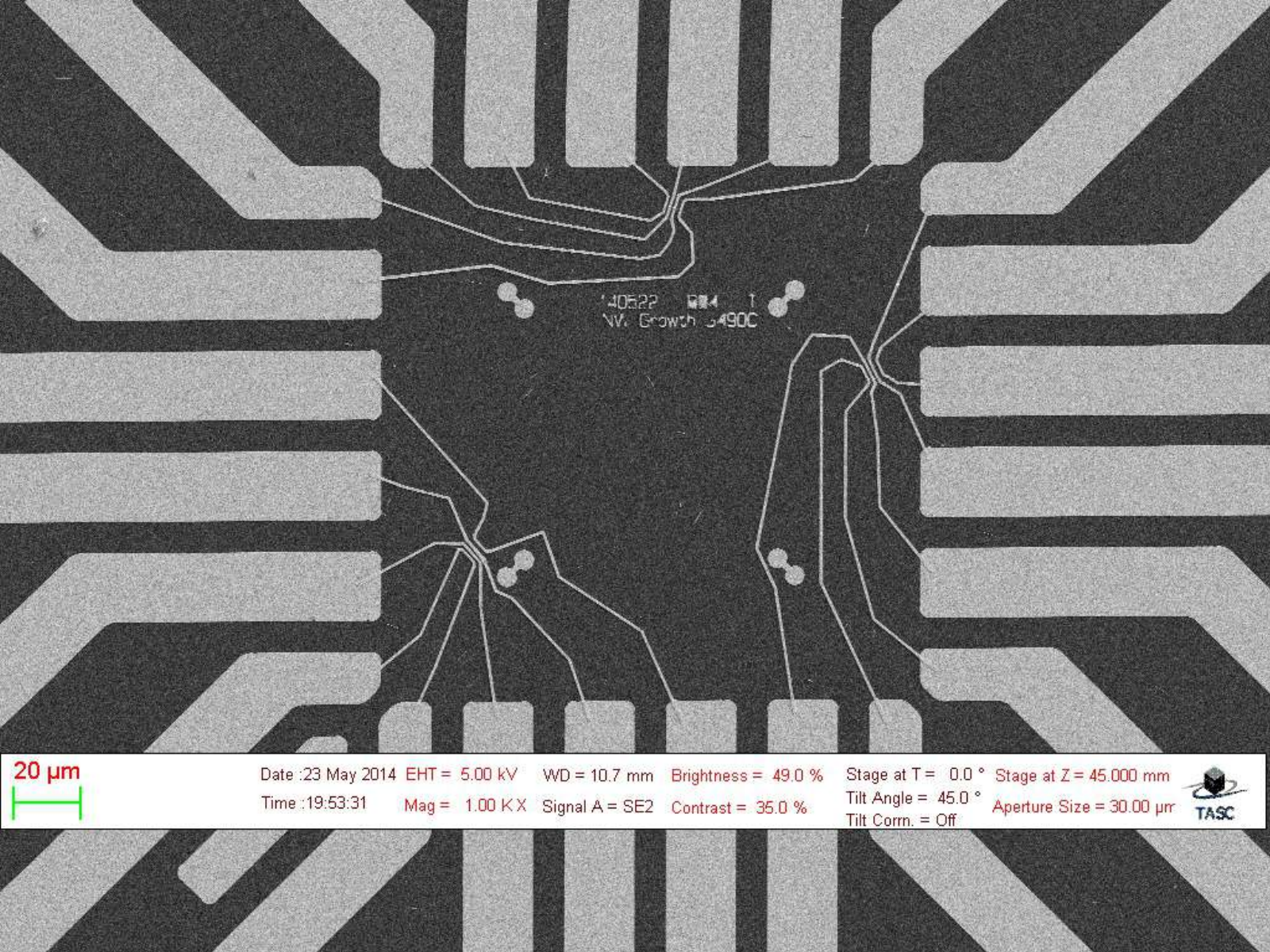}
         \caption{MEMS device}
         \label{fig:MEMS}
     \end{subfigure}
     \hspace{1em}
     \begin{subfigure}[b]{0.18\textwidth}
         \centering
         \includegraphics[width=\textwidth]{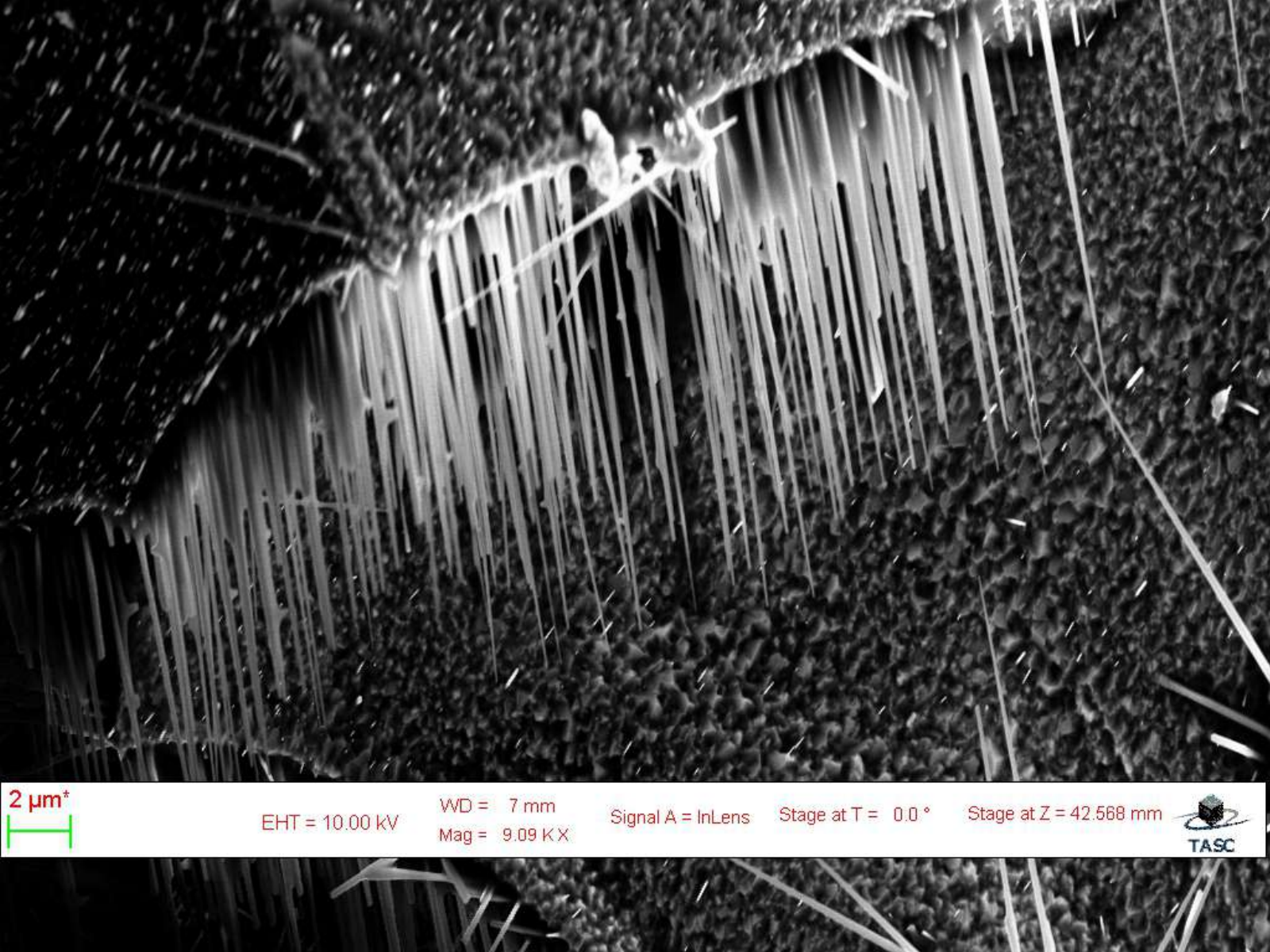}
         \caption{Nanowires}
         \label{fig:nanowires}
     \end{subfigure}
     \hspace{1em}
     \begin{subfigure}[b]{0.18\textwidth}
         \centering
         \includegraphics[width=\textwidth]{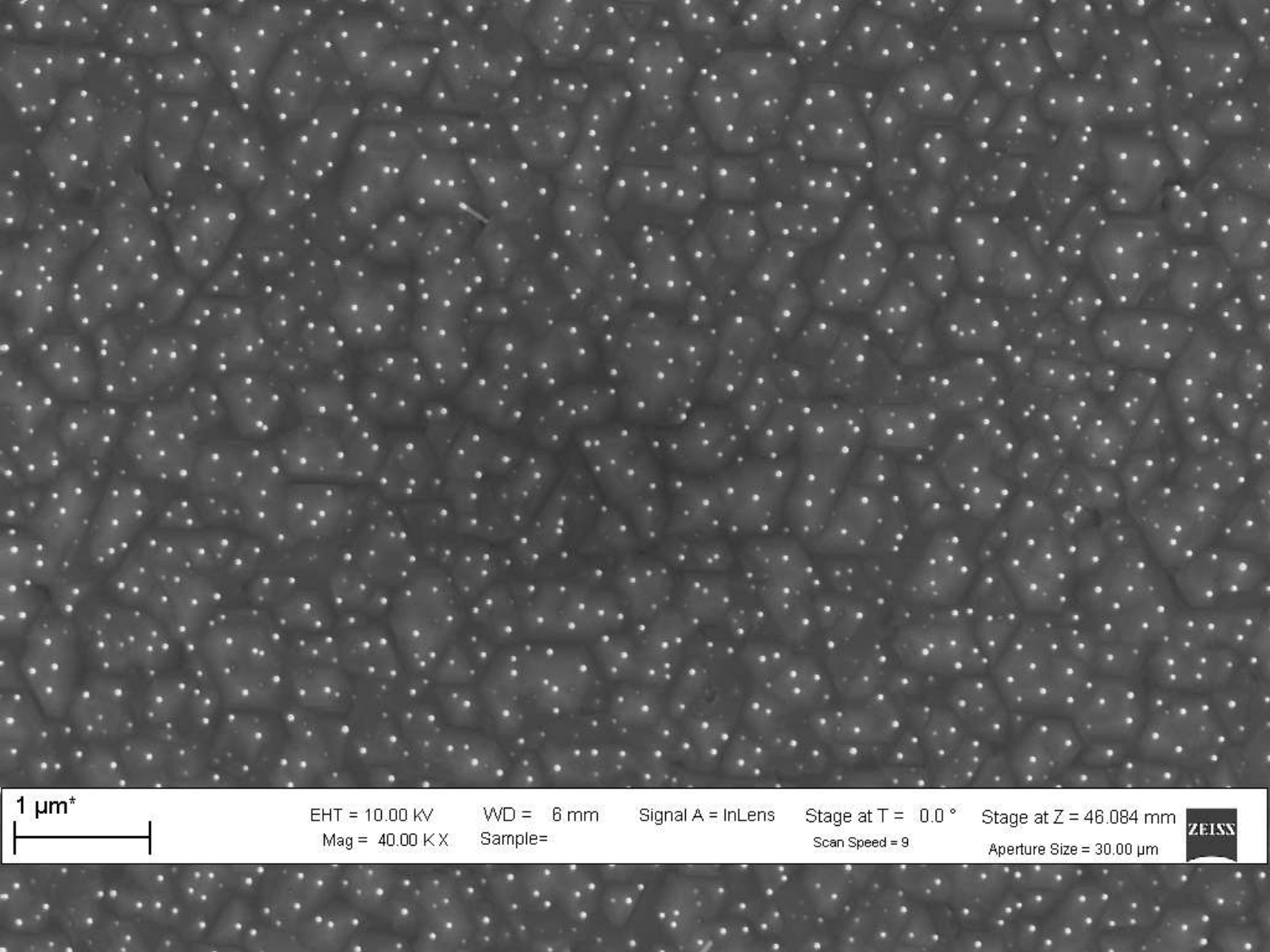}
         \caption{Particles}
         \label{fig:particles}
     \end{subfigure}
     \hspace{1em}
     \begin{subfigure}[b]{0.18\textwidth}
         \centering
         \includegraphics[width=\textwidth]{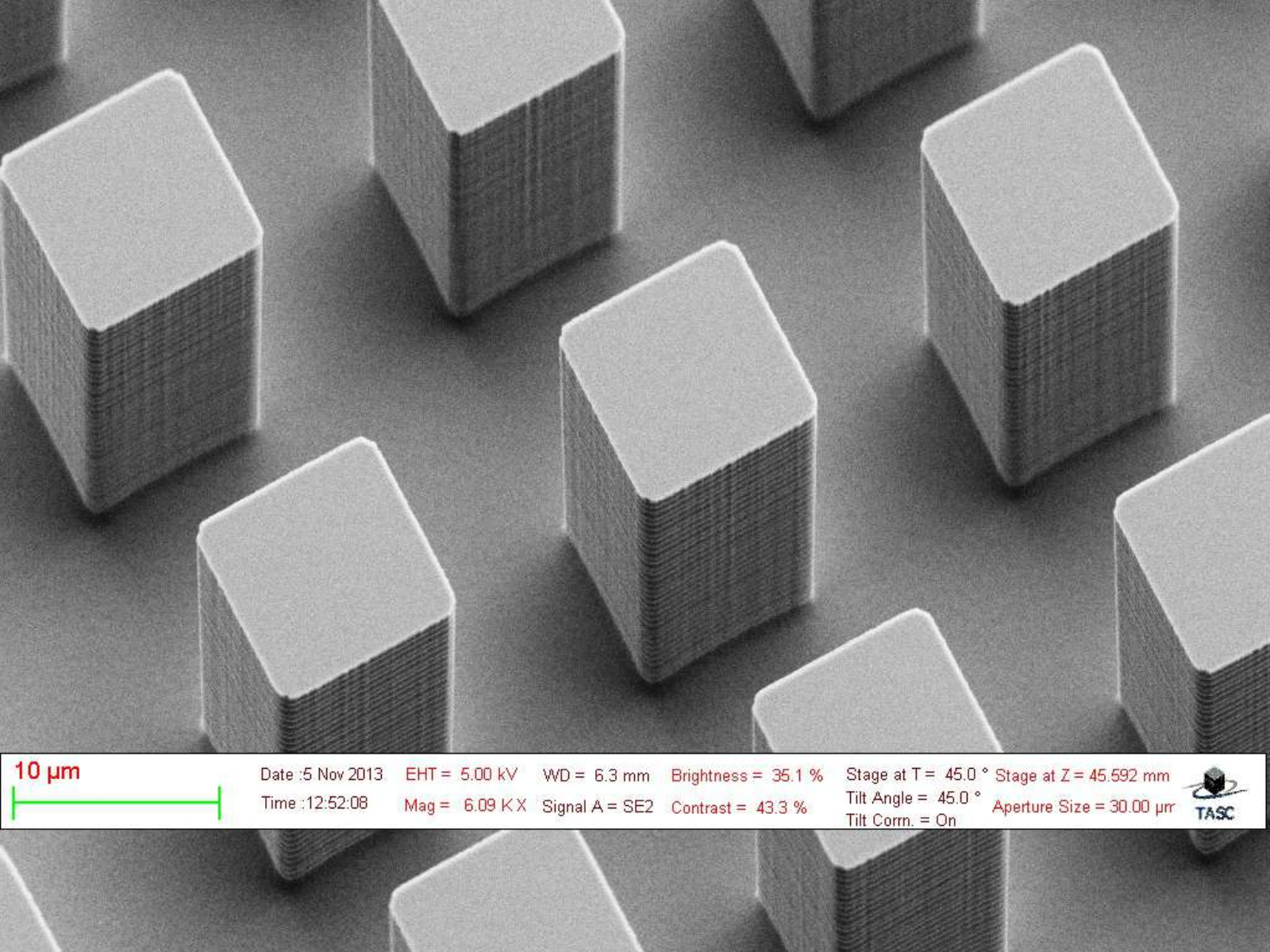}
         \caption{Pattern surface}
         \label{fig:Pattern_surface}
     \end{subfigure}
     \hspace{1em}
     \begin{subfigure}[b]{0.18\textwidth}
         \centering
         \includegraphics[width=\textwidth]{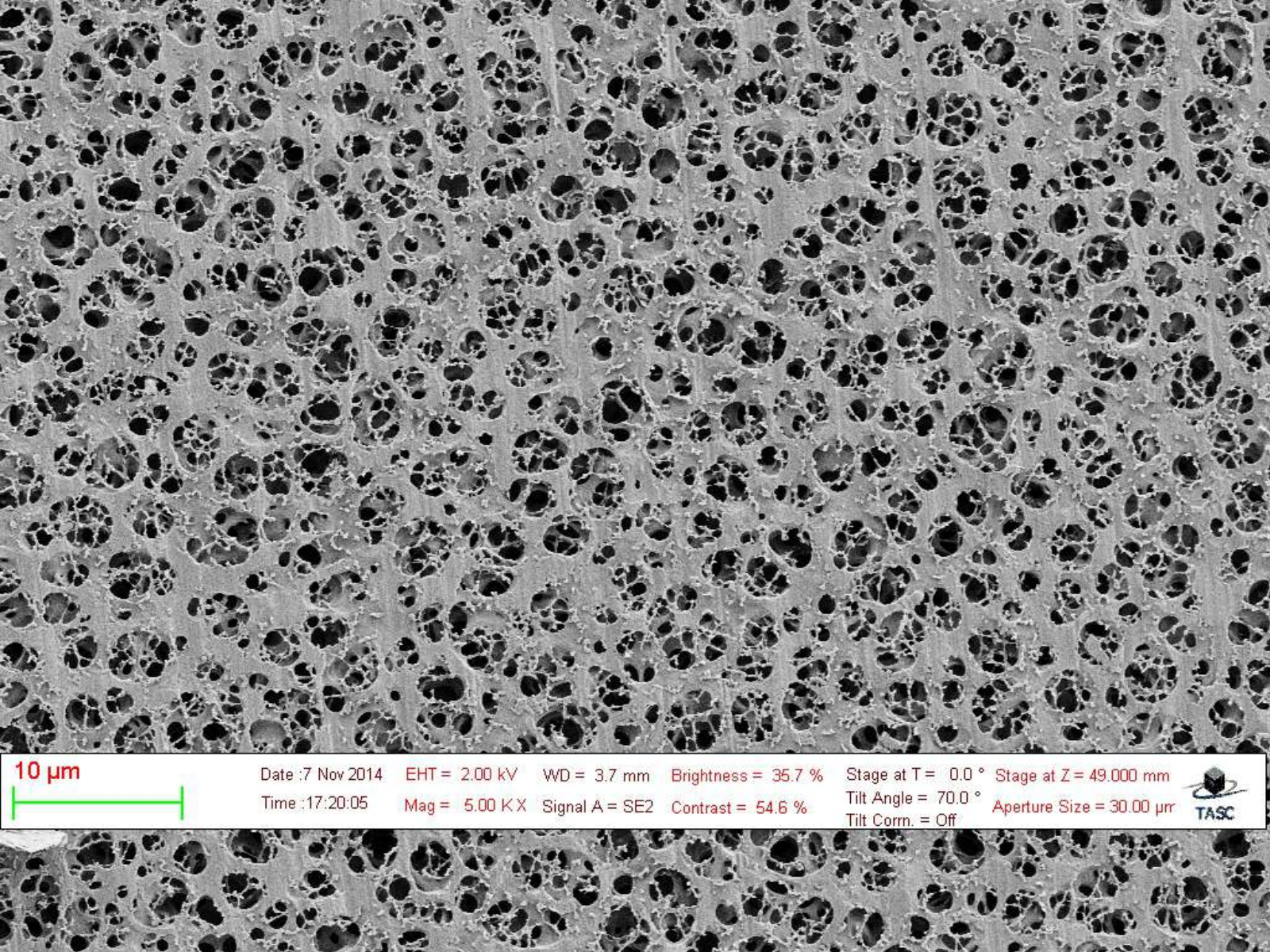}
         \caption{Porous sponge}
         \label{fig:Porous sponge}
     \end{subfigure}
     \hspace{1em}
     \begin{subfigure}[b]{0.18\textwidth}
         \centering
         \includegraphics[width=\textwidth]{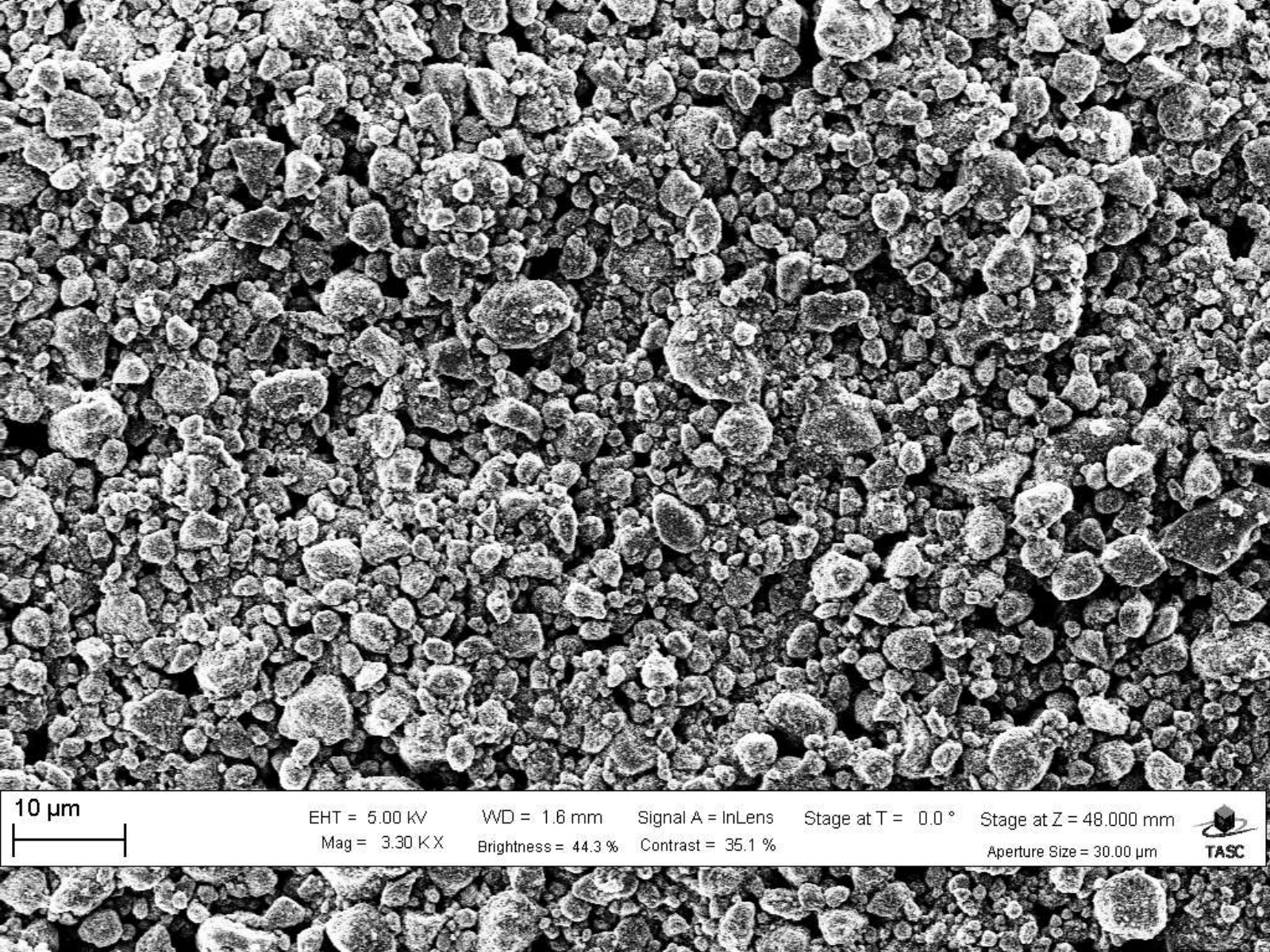}
         \caption{Powder}
         \label{fig:Powder}
     \end{subfigure}
     \hspace{1em}
     \begin{subfigure}[b]{0.18\textwidth}
         \centering
         \includegraphics[width=\textwidth]{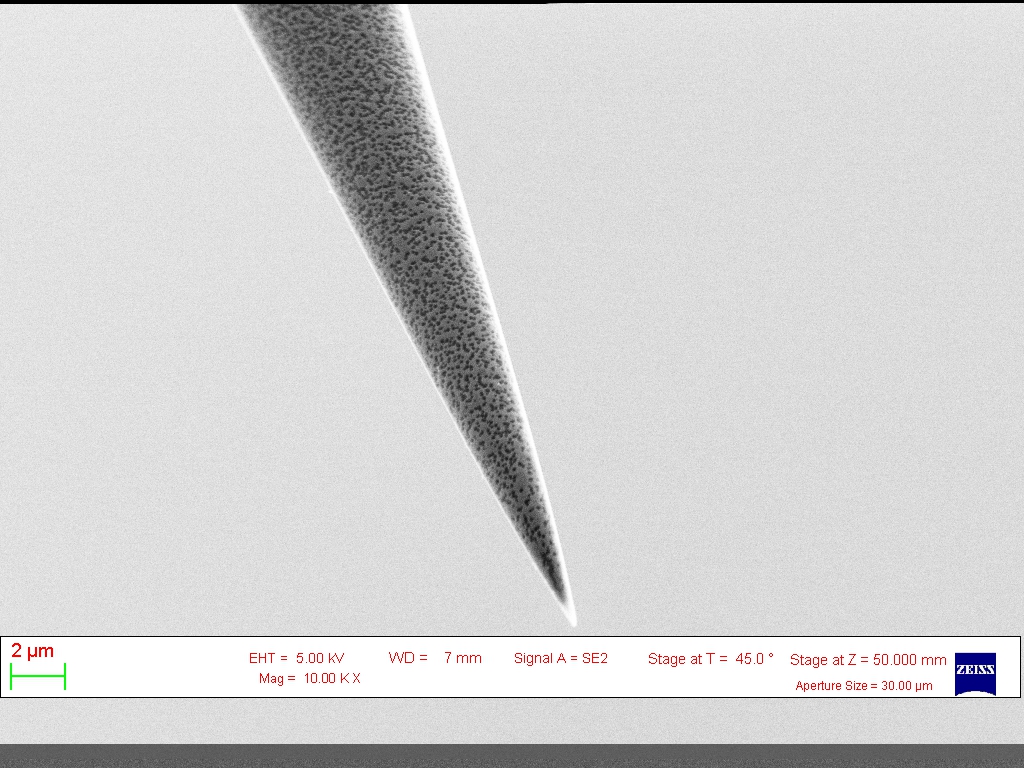}
         \caption{Tips}
         \label{fig:Tips}
     \end{subfigure}
     \vspace{-3mm}
        \caption{The SEM dataset}
        \label{fig:SEMs}
\end{figure}

In the Figure \ref{fig:SEMs} we show the representative images of the 10 different image categories in the dataset. The SEM dataset\cite{aversa2018first}, please refer to Table ~\ref{tab:SEM} has  unequal distributions in the total count of each image category in the dataset. 

\vspace{-2mm}

\vspace{0mm}
\begin{table}[htbp]
\centering
\caption{The SEM dataset: The segregation of the number of images per category is presented here.}
\vspace{-3mm}
\label{tab:SEM}
\begin{tabular}{|c|c|}
\hline
\textbf{Category}           & \textbf{Number of images} \\ \hline
Biological                  & 973                       \\
Tips                        & 1625                      \\
Fibres                      & 163                       \\
Porous Sponge               & 182                       \\
Films Coated Surface        & 327                       \\
Patterned surface           & 4756                      \\
Nanowires                   & 3821                      \\
Particles                   & 3926                      \\
MEMS devices and electrodes & 4591                      \\
Powder                      & 918                       \\ \hline
Total                       & 21282                     \\ \hline
\end{tabular}
\end{table}

\begin{figure*}[htbp]
    \centering
    \includegraphics[scale=0.55]{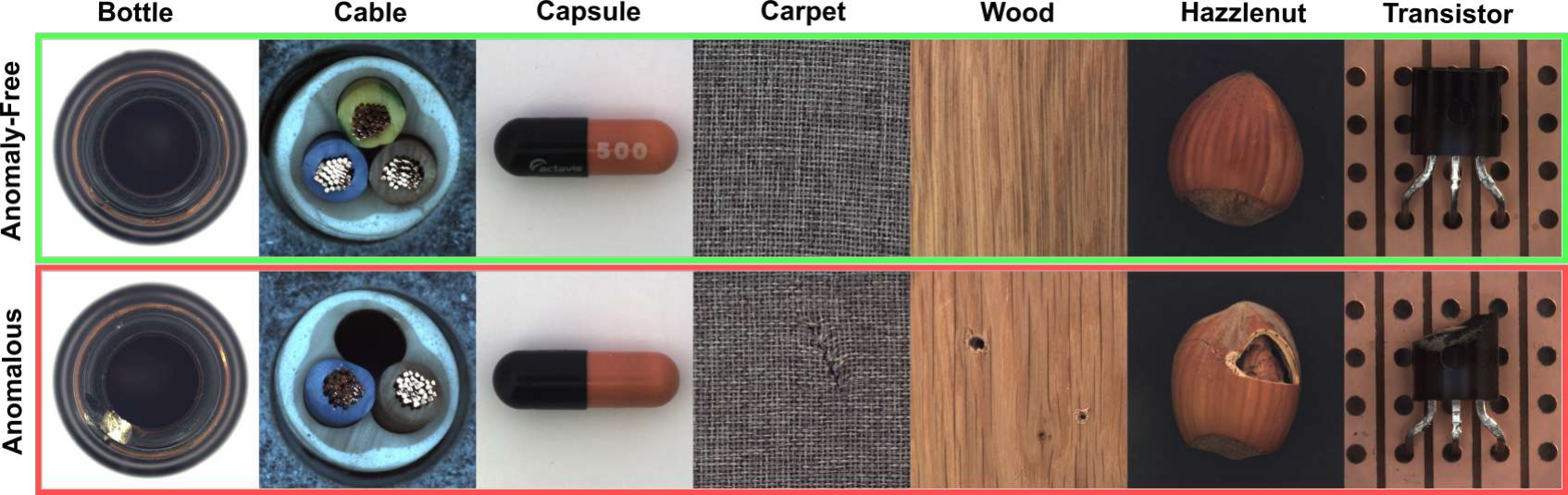}
    \vspace{-3mm}
    \caption{ Illustration of seven catogories from MVTec anomaly detection dataset.}
    \label{fig:Anomaly}
\end{figure*}
\begin{figure*}[htbp]
    \centering
    \includegraphics[scale=.55]{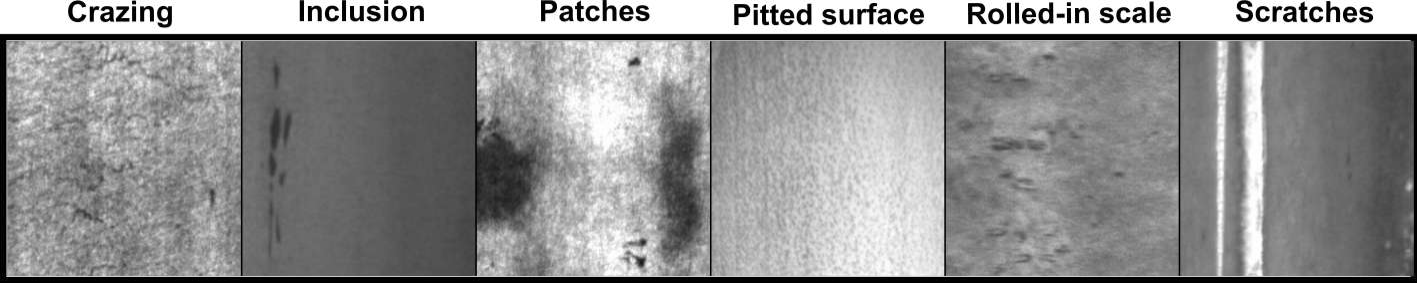}
    \vspace{-3mm}
    \caption{Illustration of six defect categories from NEU surface defect dataset of hot-rolled steel strip.}
    \label{fig:NUE}
\end{figure*}
\begin{figure*}[htbp]
    \centering
    \includegraphics[scale=0.6]{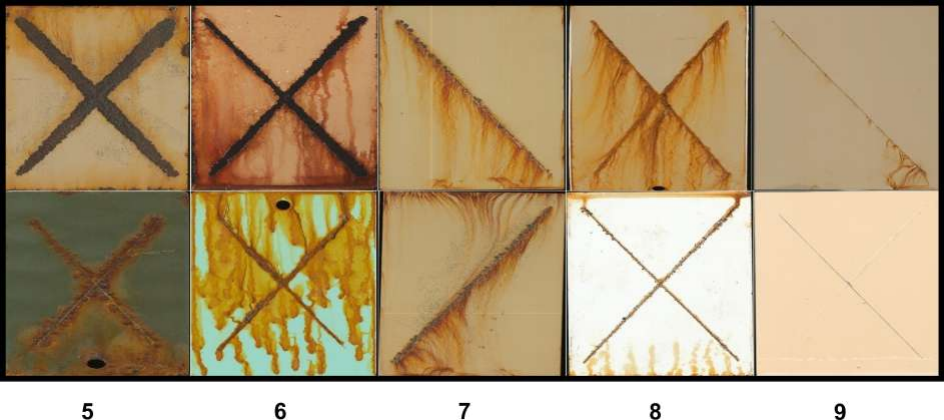}
    \vspace{-3mm}
    \caption{ Illustration of five classes of corrosion rating assigned by ASTM }
    \label{fig:corrosion}
\end{figure*}
\begin{figure*}[htbp]
    \centering
    \includegraphics[scale=0.7]{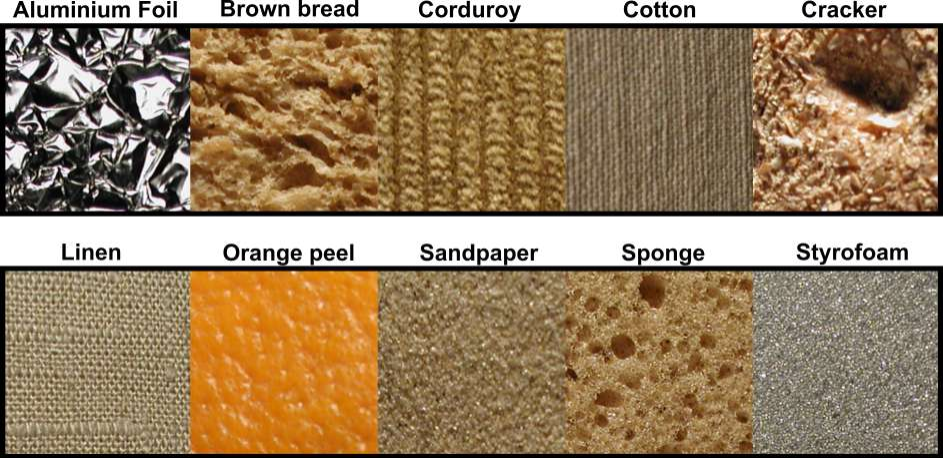}
    \vspace{-3mm}
    \caption{ Illustration of the different materials from KTH-TIPS dataset }
    \label{fig:KTH}
\end{figure*}

\vspace{-1mm}
\subsection{Hyperparameters Study}

We perform an in-depth analysis of the hyperparameters search to determine the effective model complexity of our proposed method, $\textbf{EMCNet}$. The hyperparameters of the algorithm are the dimensionality of embedding($d$), the graph-pooling ratio($p_r$), and the batch size($bs$). The hyperparameter tuning is based on grid-search technique. We learn the tuple of hyperparameters that yield the optimal performance of our proposed method on the validation set in terms of the Top-1 classification score. In each experiment, we change the hyperparameter under investigation and the remaining hyperparameters are kept constant as described in section \ref{sec:ES} to determine the effect of the hyperparameter  on the model performance. The optimal hyperparameters determined from the hyperparameter search are as follows, $d$ is 64, $p_r$ is 0.75 and $bs$ is 24.

\vspace{-3mm}
\begin{table}[ht]
\centering
\caption{Table reports the hyperparameter study}
\label{table:Hs}
\vspace{-3mm}
\begin{adjustbox}{width=1.05\columnwidth,center}
{%
\hspace{0.5cm}\begin{tabular}{@{}c|c|c|c|cc@{}}
\hline
$d$ & $32$ & $64$  & $96$ & $128$ \\
\hline
 &   0.712 $\pm$ 0.05 & \textbf{0.783 $\pm$ 0.012} &  0.789$\pm$ 0.07 &  0.791 $\pm$ 0.08 \\
 \hline
$p_r$ &  $0.25$ & $0.5$  & $0.75$ & $0.95$ \\
 \hline
  &  0.667 $\pm$ 0.08 &  0.725$\pm$ 0.03 & \textbf{ 0.783$\pm$ 0.012} &  0.748$\pm$ 0.06 \\
 \hline
$\text{bs}$ &  $12$ & $24$  & $48$ & $64$ \\
 \hline  
  & 0.706 $\pm$ 0.04 & \textbf{0.783 $\pm$ 0.012}  &  0.790$\pm$ 0.07 &  0.786$\pm$ 0.05  \\
\hline
\end{tabular}
}
\end{adjustbox}

\end{table}

\vspace{-4mm}
\subsection{Benchmarking on Material datasets}
\begin{itemize}
    \item \textbf{ MVTec dataset\footnote{Datasource: \url{https://www.mvtec.com/company/research/datasets.}}}\cite{bergmann2021mvtec} is an open-sourced anomaly detection benchmark dataset of industrially driven products. The database contains $\approx$5000 high-resolution RGB images. The images are classified into 15 different categories of objects(e.g \textit{bottle, cable, metal nut, hazelnut, toothbrush, capsule, pills, screw, transistor, zipper}) and textures(e.g \textit{carpet, grid, leather, tile, and wood}). The training/validation set contains 3629 images. The test set contains the remaining images. The training set contains normal data(anomaly-free), whereas the test set contains both the normal and the anomalies(consists of different kinds of defects). A few representative images of a sample of defect categories, along with the corresponding defect-free images are shown in Figure \ref{fig:Anomaly}. The dataset is highly class imbalanced.  We evaluate the performance of our proposed method on the supervised anomaly detection task(binary classification) in comparison with several baseline algorithms.
    \item \textbf{NEU surface defect dataset(NEU-SDD) \footnote{Datasource: \url{http://faculty.neu.edu.cn/yunhyan/NEU_surface_defect_database.html}}}\cite{deshpande2020one} contains 1800 gray-scale images of hot-rolled steel strip. The database is classified into six classes of surface defects that includes \textit{crazing, inclusion, patches, pitted surface, rolled-in scale, and scratches}. Each class has 300 sample images. A few representative images of the surface defects are shown in Figure \ref{fig:NUE}. We evaluate the performance of our proposed method on the multiclass classification task on the balanced dataset in comparison with several baseline algorithms.
    \item \textbf{Corrosion Image Dataset(CMI)\footnote{\url{https://arl.wpi.edu/corrosion_dataset}}} is utilized to benchmark various algorithmic approaches for automation of corrosion assessment in materials. The dataset contains 600 images of corroding panels of resolution 512$\times$512 pixels. Each panel in the dataset is annotated by the experts to provide the measure of the corrosion through the ASTM standards. In general, the corrosion rating  is assigned on a scale of discrete integers from 1 to 10. The corrosion rating of 10 implies that the panel is in the initial stage of corrosion. The data presenters projects it as a class-balanced dataset by rejecting the panels with high corrosion ratings(1-4). Thus the dataset is classified into five corrosion ratings(5-9). Each corrosion rating based image category has 120 images. The sample images per corrosion rating are shown in Figure\ref{fig:corrosion}. We evaluate the performance of our proposed method on the multi-class classification task in comparison with several baseline algorithms.      
    \item \textbf{KTH-TIPS\footnote{\url{https://www.csc.kth.se/cvap/databases/kth-tips/index.html}}} is a texture database which contains images of varying illumination, pose and scale of ten different materials. It contains a sample of 810 images belonging to different classes of materials such as \textit{crumpled aluminum foil, brown bread, corduroy, cotton, cracker, linen, orange peel, sandpaper, sponge, and styrofoam}. A few example images per category are shown in Figure\ref{fig:KTH}. The image resolution in the dataset is 200$\times$200 pixels. We evaluate the performance of our proposed method on the multi-class classification task in comparison with several baseline algorithms.      
\end{itemize}

We report the results of the baseline algorithms and our model performance in Table ~\ref{table:PD} on the open-sourced benchmark datasets. The evaluation metric is the 
Top-1 classification accuracy. We achieve SOTA performance on all the datasets. 

\vspace{-2mm}
\begin{table}[htbp]
\centering
\setlength{\tabcolsep}{3pt}
\caption{Performance comparison on the datasets.}
\label{table:PD}
\vspace{-4mm}
\begin{tabular}{cc|ccccc}
\hline
\multicolumn{2}{c|}{\textbf{Algorithms}}                                      & \textbf{MVTec} & \textbf{NEU-SDD} & \textbf{CMI} & \textbf{KTH-TIPS}  \\ \hline
\multicolumn{1}{c|}{\multirow{4}{*}{\rotatebox[origin=c]{90}{\textbf{Baselines}}}} & ResNet       & 0.912             & 0.926             & 0.911             & 0.932             &             \\
\multicolumn{1}{c|}{}                                          & GoogleNet    & 0.926             & 0.933             & 0.913             & 0.924              \\
\multicolumn{1}{c|}{}                                          & SqueezeNet   & 0.931             & 0.947             & 0.937             & 0.958              \\ 
\multicolumn{1}{c|}{}                                          & VanillaViT    & 0.954             & 0.953             & 0.955             & 0.966 \\ 
\hline
\multicolumn{1}{c|}{}                                          & \textbf{EMCNet}     &    \textbf{0.983}               &     \textbf{0.978}              &      \textbf{0.971}             &    \textbf{0.992}               &                     \\ \hline
\end{tabular}
\end{table}

\end{document}